\documentclass{article}

\PassOptionsToPackage{numbers}{natbib}
\usepackage[final]{neurips_2020}

\usepackage[utf8]{inputenc} 
\usepackage[T1]{fontenc}    
\usepackage{hyperref}       
\usepackage{url}            
\usepackage{booktabs}       
\usepackage{amsfonts}       
\usepackage{nicefrac}       
\usepackage{microtype}      

\usepackage{our_style}

\usepackage{color}

\title{Kernelized information bottleneck leads to biologically plausible 3-factor Hebbian learning in deep networks}

\author{%
  Roman Pogodin \\
  Gatsby Computational Neuroscience Unit\\
  University College London\\
  London, W1T 4JG \\
  \texttt{roman.pogodin.17@ucl.ac.uk} \\
  \And
  Peter E. Latham \\
  Gatsby Computational Neuroscience Unit\\
  University College London\\
  London, W1T 4JG \\
  \texttt{pel@gatsby.ucl.ac.uk} \\
}

\begin{document}

\maketitle

\begin{abstract}
  
  
  The state-of-the art machine learning approach to training deep neural networks, backpropagation, is implausible for real neural networks: neurons need to know their outgoing weights; training alternates between a bottom-up forward pass (computation) and a top-down backward pass (learning); and the algorithm often needs precise labels of many data points. Biologically plausible approximations to backpropagation, such as feedback alignment, solve the weight transport problem, but not the other two. Thus, fully biologically plausible learning rules have so far remained elusive. Here we present a family of learning rules that does not suffer from any of these problems. It is motivated by the information bottleneck principle (extended with kernel methods), in which networks learn to compress the input as much as possible without sacrificing prediction of the output. The resulting rules have a 3-factor Hebbian structure: they require pre- and post-synaptic firing rates and an error signal -- the third factor -- consisting of a global teaching signal and a layer-specific term, both available without a top-down pass. They do not require precise labels; instead, they rely on the similarity between pairs of desired outputs. Moreover, to obtain good performance on hard problems and retain biological plausibility, our rules need divisive normalization -- a known feature of biological networks. Finally, simulations show that our rules perform nearly as well as backpropagation on image classification tasks.
\end{abstract}

\section{Introduction}

\begin{figure}[t]
    \centering
    \includegraphics[width=\textwidth]{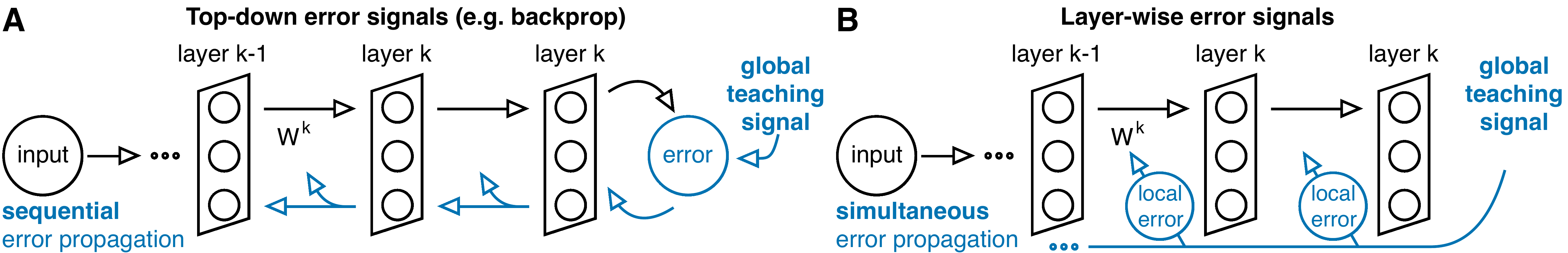}
    \caption{\textbf{A.} The global error signal is propagated to each layer from the layer above, and used to update the weights. \textbf{B.} The global error signal is sent directly to each layer.}
    \label{fig:top_down_vs_layer}
\end{figure}

Supervised learning in deep networks is typically done using the backpropagation algorithm (or backprop), but in its present form it cannot explain learning in the brain \cite{richards2019deep}. There are three reasons for this: weight updates require neurons to know their \textit{outgoing} weights, which they do not (the weight transport problem); the forward pass for computation and the backward pass for weight updates need separate pathways and have to happen sequentially (preventing updates of earlier layers before the error is propagated back from the top ones, see \cref{fig:top_down_vs_layer}A); and a large amount of precisely labeled data is needed.

While approximations to backprop such as feedback alignment \cite{akrout2019deep,lillicrap2016random} can solve the weight transport problem, they do not eliminate the requirement for a backward pass or the need for labels. There have been suggestions that a backward pass could be implemented with apical dendrites \cite{sacramento2018dendritic}, but it's not clear how well the approach scales to large networks, and the backward pass still has to follow the forward pass in time.

Backprop is not, however, the only way to train deep feedforward networks.
An alternative is to use so-called layer-wise update rules, which require only activity in adjacent (and thus connected) layers, along with a global error signal (\cref{fig:top_down_vs_layer}B). Layer-wise training removes the need for both weight transport and a backward pass, and there is growing evidence that such an approach can work as well as backprop \cite{pmlr-v97-belilovsky19a,nokland2019training, lowe2019putting}. However, while such learning rules are local in the sense that they mainly require activity only in adjacent layers, that does not automatically imply biological plausibility.

Our work focuses on finding a layer-wise learning rule that is biologically plausible. For that we take inspiration from the information bottleneck principle \cite{tishby2000information, shwartz2017opening}, in which every layer minimizes the mutual information between its own activity and the input to the network, while maximizing the mutual information between the activity and the correct output (e.g., a label). Estimating the mutual information is hard \cite{alemi2016deep}, so \cite{ma2019hsic} proposed the HSIC bottleneck: instead of mutual information they used ``kernelized cross-covariance'' called Hilbert-Schmidt independence criterion (HSIC). HSIC was originally proposed in as a way to measure independence between distributions \cite{gretton2005measuring}. Unlike mutual information, HSIC is easy to estimate from data \cite{gretton2005measuring}, and the information bottleneck objective keeps its intuitive interpretation. Moreover, as we will see, for classification with roughly balanced classes it needs only pairwise similarities between labels, which results in a binary teaching signal.

Here we use HSIC, but to achieve biologically plausible learning rules we modify it in two ways: we replace the HSIC between the input and activity with the kernelized covariance, and we approximate HSIC with ``plausible HSIC'', or pHSIC, the latter so that neurons don't need to remember their activity over many data points. (However, the objective function becomes an upper bound to the HSIC objective.) The resulting learning rules have a 3-factor Hebbian structure: the updates are proportional to the pre- and post-synaptic activity, and are modulated by a third factor (which could be a neuromodulator \cite{gerstner2018eligibility}) specific to each layer. 
In addition, to work on hard problems and remain biologically plausible, our update rules need divisive normalization, a computation done by the primary visual cortex and beyond \cite{carandini2012normalization, olsen2010divisive}.

In experiments we show that plausible rules generated by pHSIC work nearly as well as backprop on MNIST \cite{lecun2010mnist}, fashion-MNIST \cite{xiao2017fmnist}, Kuzushiji-MNIST \cite{clanuwat2018kmnist} and CIFAR10 \cite{krizhevsky2009cifar} datasets. This significantly improves the results from the original HSIC bottleneck paper \cite{ma2019hsic}.

\section{Related work}
Biologically plausible approximations to backprop solve the weight transport problem in multiple ways. Feedback alignment (FA) \cite{lillicrap2016random} and direct feedback alignment (DFA) \cite{nokland2016direct} use random fixed weights for the backward pass but scale poorly to hard tasks such as CIFAR10 and ImageNet \cite{moskovitz2018feedback, bartunov2018assessing}, However, training the feedback pathway to match the forward weights can achieve backprop-level performance \cite{akrout2019deep}. The sign symmetry method \cite{liao2015important} uses the signs of the feedforward weights for feedback and therefore doesn't completely eliminate the weight transport problem, but it scales much better than FA and DFA \cite{akrout2019deep, xiao2018biologically}. Other methods include target prop \cite{bengio2014auto, lee2015difference} (scales worse than FA \cite{bartunov2018assessing}) and equilibrium prop \cite{scellier2017equilibrium} (only works on simple tasks, but can work on CIFAR10 at the expense of a more complicated learning scheme \cite{laborieux2020scaling}). However, these approaches still need to alternate between forward and backward passes.

To avoid alternating forward and backward passes, layer-wise objectives can be used. A common approach is layer-wise classification: \cite{mostafa2018deep} used fixed readout weights in each layer (leading to slightly worse performance than backprop); \cite{nokland2019training} achieved backprop-like performance with trainable readout weights on CIFAR10 and CIFAR100; and \cite{pmlr-v97-belilovsky19a} achieved backprop-like performance on ImageNet with multiple readout layers. However, layer-wise classification needs precise labels and local backprop (or its approximations) for training. 
Methods such as contrastive learning \cite{lowe2019putting} and information \cite{shwartz2017opening} or HSIC \cite{ma2019hsic} bottleneck and gated linear networks \cite{veness2017online, veness2019gated} provide alternatives to layer-wise classification, but don't focus on biological plausibility.
Biologically plausible  alternatives with weaker supervision include similarity matching \cite{nokland2019training, qin2020supervised}, with \cite{nokland2019training} reporting a backprop-comparable performance using cosine similarity, and fully unsupervised rules such as \cite{pehlevan2018similarity, krotov2019unsupervised}. Our method is related to similarity matching; see below for additional discussion.

\section{Training deep networks with layer-wise objectives: a kernel methods approach and its plausible approximation}
\label{sec:method}
Consider an $L$-layer feedforward network with input $\bb x$, layer activity $\bb z^k$ (for now without divisive normalization) and output $\hat{\bb y}$,
\begin{equation}
    \bb z^1 = f\brackets{\bb W^1 \bb x},\dots,\,\bb z^L = f\brackets{\bb W^L \bb z^{L-1}};\ \hat{\bb y} = f\brackets{\bb W^{L+1}\bb z^L}\,.
    \label{eq:feedforward_net}
\end{equation}
The standard training approach is to minimize a loss, $l(\bb y, \hat{\bb y})$, with respect to the weights, where $\bb y$ is the desired output and $\hat{\bb y}$ is the prediction of the network.
Here, though, we take an alternative approach: we use layer-wise objective functions, $l_k(\bb x, \bb z^k, \bb y)$ in layer $k$, and minimize each $l_k(\bb x, \bb z^k, \bb y)$ with respect to the weight in that layer, $\bb W^k$ (simultaneously for every $k$).
The performance of the network is still measured with respect to $l(\bb y, \hat{\bb y})$, but that quantity is explicitly minimized only with respect to the output weights, $\bb W^{L+1}$.

To choose the layer-wise objective function, we turn to the information bottleneck \cite{tishby2000information}, which minimizes the information between the input and the activity in layer $k$, while maximizing the information between the activity in layer $k$ and the output \cite{shwartz2017opening}.
Information, however, is notoriously hard to compute \cite{alemi2016deep}, and so \cite{ma2019hsic}  proposed an alternative based on the Hilbert-Schmidt Independence Criterion (HSIC) -- the HSIC bottleneck. HSIC is a kernel-based method for measuring independence between probability distribution \cite{gretton2005measuring}. Similarly to the information bottleneck, this method tries to balance compression of the input with prediction of the correct output, with a (positive) balance parameter $\gamma$,

\begin{equation}
    \min_{\bb W^k} \hsic{X}{Z^k}-\gamma\, \hsic{Y}{Z^k},\ k=1,\dots,L\,,
    \label{eq:ma_objective}
\end{equation}
where $X,Z^k$ and $Y$ are random variables, with a distribution induced by the input (the $\bb x$) and output (the $\bb y$). HSIC is a measure of dependence: it is zero if its arguments are independent, and increases as dependence increases,
\begin{align}
    \hsic{A}{B} \,&= \int\Delta P_{ab}(\bb a_1, \bb b_1) \,k(\bb a_1, \bb a_2) \,k(\bb b_1,\bb b_2)\, \Delta P_{ab}(\bb a_2, \bb b_2)\,,
    \label{eq:hsic}
\end{align}
where
\begin{align}
    \Delta P_{ab}(\bb a, \bb b) \,& \equiv (p_{ab}(\bb a, \bb b)-p_a(\bb a)p_b(\bb b) )\,d\bb a\, d\bb b \, .
\end{align}
The kernels, $k(\cdot,\cdot)$ (which might be different for $\bb a$ and $\bb b$), are symmetric and positive definite functions, the latter to insure that HSIC is non-negative. More details on kernels and HSIC are given in \cref{app:sec:kernels}. 

HSIC gives us a layer-wise cost function, which eliminates the need for backprop.
However, there is a downside: estimating it from data requires memory.
This becomes clear when we consider the empirical estimator of \cref{eq:hsic} given $m$ observations \cite{gretton2005measuring},
\begin{equation}
\begin{split}
    \widehat{\mathrm{HSIC}}(A,B) \!& =\!
    \frac{1}{m^2}\!\sum_{ij}\! k(\bb a_i, \bb a_j) k(\bb b_i, \bb b_j)
    +
    \frac{1}{m^2}\!\sum_{ij}\! k(\bb a_i, \bb a_j)
    \frac{1}{m^2}\!\sum_{kl}\! k(\bb b_k, \bb b_l)
    \\ &
    - \!
    \frac{2}{m^3}\!\sum_{ijk} k(\bb a_i, \bb a_k)k(\bb b_j, \bb b_k)
    \, .
    \label{eq:hsic_bar}
\end{split}
\end{equation}
In a realistic network, data points are seen one at a time, so to compute the right hand side from $m$ samples, $m-1$ data point would have to be remembered.
We solve this in the usual way, by stochastic gradient descent. For the first term we use two data points that are adjacent in time; for the second, we accumulate and store the average over the kernels (see \cref{eq:empirical_centering} below). The third term, however, is problematic; to compute it, we would have to use three data points. Because this is implausible, we make the approximation
\begin{align}
    \frac{1}{m}\!\sum_{k} k(\bb a_i, \bb a_k)k(\bb b_j, \bb b_k)
    \approx
    \frac{1}{m^2}\!\sum_{kl} k(\bb a_i, \bb a_k)k(\bb b_j, \bb b_l)
    \, .
\end{align}
Essentially, we replace the third term in Eq.~\eqref{eq:hsic_bar} with the second. This leads to ``plausible'' HSIC, which we call pHSIC,
\begin{align}
    \mathrm{p}\hsic{A}{B} \,&=    \brackets{\expect_{\bb a_1\bb b_1}\expect_{\bb a_2\bb b_2} - \expect_{\bb a_1}\expect_{\bb b_1}\expect_{\bb a_2}\expect_{\bb b_2}}\brackets{k(\bb a_1, \bb a_2) \,k(\bb b_1,\bb b_2)}
    \, .
    \label{eq:hsic_plausible}
\end{align}

While pHSIC is easier to compute than HSIC, there is still a potential problem: computing $\hsic{X}{Z^k}$ requires $k(\bb x_i, \bb x_j)$, as can be seen in the above equation. But if we knew how to build a kernel that gives a reasonable distance between inputs, we wouldn't have to train the network for classification.
So we make one more change: rather than minimizing the dependence between $X$ and $Z^k$ (by minimizing the first term in \cref{eq:ma_objective}), we minimize the (kernelized) covariance of $Z^k$. 
To do this, we replace $X$ with $Z^k$ in Eq.~\eqref{eq:ma_objective}, and define $\mathrm{p}\hsic{A}{A}$ via Eq.~\eqref{eq:hsic_plausible} but with $p_{ab}(\bb a, \bb b)$ set to $p_a(\bb a)\delta(\bb a - \bb b)$,
\begin{align}
\mathrm{p}\hsic{A}{A} =
\expect_{\bb a_1}\expect_{\bb a_2} \brackets{k(\bb a_1, \bb a_2) }^2 -
\brackets{
\expect_{\bb a_1}\expect_{\bb a_2} k(\bb a_1, \bb a_2) }^2\, .
\label{eq:phsic_aa}
\end{align}
This gives us the new objective,
\begin{equation}
    \min_{\bb W^k} \mathrm{p}\hsic{Z^k}{Z^k}-\gamma \, \mathrm{p}\hsic{Y}{Z^k},\ k=1,\dots,L\,.
    \label{eq:zy_objective_plausible}
\end{equation}
The new objective preserves the intuition behind the information bottleneck, which is to throw away as much information as possible about the input. It's also an upper bound on the ``true'' HSIC objective as long as the kernel over the desired outputs is centered ($\expect_{\bb y_1}k(\bb y_1, \bb y_2)=0$; see \cref{app:subsec:phsic}), and doesn't significantly change the performance (see \cref{app:sec:experiments}).

Centering the output kernel is straightforward  in classification with balanced classes: take $\bb y$ to be centered one-hot encoded labels (for $n$ classes, $y_i=1-1/n$ for label $i$ and $-1/n$ otherwise), and use the cosine similarity kernel $k(\bb y_1,\bb y_2)=\bb y_1\trans\bb y_2 / \|\bb y_1\|\|\bb y_2\|$. In addition, the teaching signal is binary in this case: $k(\bb y_i, \bb y_j)=1$ if $\bb y_i =\bb y_j$ and  $-1/(n-1)$ otherwise. We will use exactly this signal in our experiments, as the datasets we used are balanced. For slightly unbalanced classes, the signal can still be binary (see \cref{app:subsec:binary_teaching}).

When $\gamma=2$, our objective is related to similarity matching. For the cosine similarity kernel over activity, it is close to \cite{nokland2019training}, and for the linear kernel, it is close to \cite{qin2020supervised, pehlevan2018similarity}. However, the update rule for the cosine similarity kernel is implausible, and for the linear kernel the rule performs poorly (see below and in \cref{app:sec:update_derivations}). We thus turn to the Gaussian kernel.

\section{Circuit-level details of the gradient: a hidden 3-factor Hebbian structure}
\subsection{General update rule}
To derive the update rule for gradient descent, we need to estimate the gradient of \cref{eq:zy_objective_plausible}. This is relatively straightforward, so we leave the derivation to \cref{app:sec:update_derivations}, and just report the update rule,
\begin{align}
    \Delta \bb W^k  &\,\propto \sum_{ij} \brackets{ \gamma \boldring k(\bb y_i, \bb y_j) - 2 \boldring k(\bb z^{k}_i,\bb z^{k}_j)} \deriv{}{\bb W^k} k(\bb z^{k}_i,\bb z^{k}_j)\,,\label{eq:generic_update_rule}
\end{align}
where the circle above $k$ means empirical centering,
\begin{equation}
     \boldring k(\bb a_i, \bb a_j) \equiv k(\bb a_i, \bb a_j) - \frac{1}{m^2}\sum_{i'j'}{k(\bb a_{i'}, \bb a_{j'})}\,.
    \label{eq:empirical_centering}
\end{equation}

\begin{figure}[h]
    \centering
    \includegraphics[width=\textwidth]{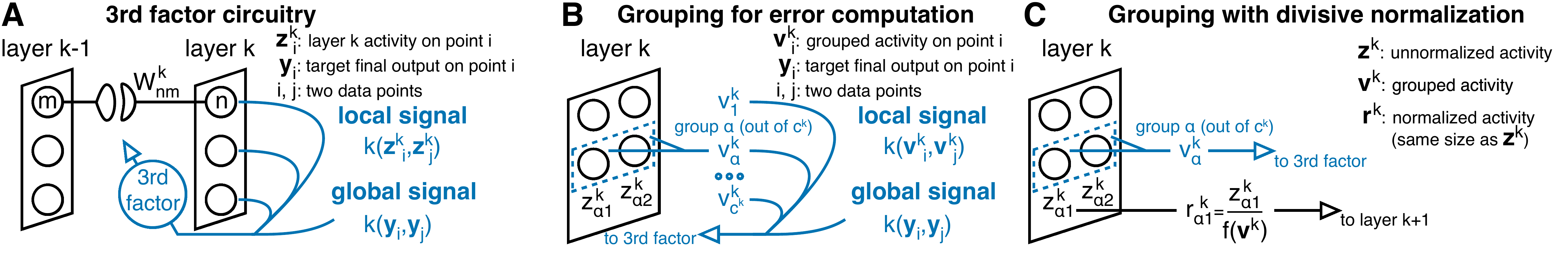}
    \caption{\textbf{A.} The weight update uses a 3rd factor consisting of a global teaching signal and a local (to the layer) signal, both capturing similarities between two data points. \textbf{B.} Grouping: neurons in layer $k$ form $c^k$ groups, each represented by a single number; those numbers are used to compute the local signal. \textbf{C.} Grouping with divisive normalization: the activity of every neuron is normalized using the grouped signal before passing to the next layer (the error is computed as in B).}
    \label{fig:circuitry}
\end{figure}

This rule has a 3-factor form with a global and a local part (\cref{fig:circuitry}A): for every pair of points $i$ and $j$, every synapse needs the same multiplier, and this multiplier needs the similarities between labels (the global signal) and layer activities (the local signal) on two data points. As mentioned in the previous section, on our problems the teaching signal $k(\bb y_i, \bb y_j)$ is binary, but we will keep the more general notation here.

However, the derivative in \cref{eq:generic_update_rule} is not obviously Hebbian. In fact, it gives rise to a simple Hebbian update only for some kernels. Below we consider the Gaussian kernel, and in \cref{app:subsec:cossim_derivation} we show that the cosine similarity kernel (used in \cite{nokland2019training}) produces an unrealistic rule.

\subsection{Gaussian kernel: two-point update}

The Gaussian kernel is given by $k(\bb z^{k}_i,\bb z^{k}_j)=\exp(-\|\bb z^{k}_i-\bb z^{k}_j\|^2/2\sigma^2)$. To  compute updates from a stream of data (rather than batches), we approximate  the sum in \cref{eq:generic_update_rule} with only two points, for which we'll again use $i$ and $j$. If we take a linear fully connected layer (see \cref{app:subsec:gauss_derivation} for the general case), the update over two points is
\begin{subequations}
\begin{align}
    \Delta \bb W^k  &\,\propto M^k_{ij} (\bb z^{k}_i - \bb z^{k}_{j}) (\bb z^{k-1}_i - \bb z^{k-1}_{j})\trans\\
    M^k_{ij} \,&= -\frac{1}{\sigma^2}\brackets{ \gamma\,\boldring{k}(\bb y_i, \bb y_{j}) -2\, \boldring{k}(\bb z^{k}_i,\bb z^{k}_{j})}\,k(\bb z^{k}_i,\bb z^{k}_{j})\, .
\end{align}\label{eq:gauss_update_ij}
\end{subequations}
\!\!Here $M^k_{ij}$ is the layer-specific third factor.

The role of the third factor is to ensure that if labels, $\bb y_i$ and $\bb y_j$, are similar, then the activity, $\bb z_i^k$ and $\bb z_j^k$, is also similar, and vice-versa.
To see why it has that effect, assume $\bb y_i$ and $\bb y_j$ are similar and $\bb z_i^k$ and $\bb z_j^k$ are not. That makes $M_{ij}^k$ negative, so the update rule is anti-Hebbian, which tends to move activity closer together.
Similarly, if $\bb y_i$ and $\bb y_j$ are not similar and $\bb z_i^k$ and $\bb z_j^k$ are, $M_{ij}^k$ is positive, and the update rule is Hebbian, which tends to move activity farther apart.

\subsection{Gaussian kernel with grouping and divisive normalization}
\label{subsec:divnorm}
The Gaussian kernel described above works well on small problems (as we'll see below), but in wide networks it needs to compare very high-dimensional vectors, for which there is no easy metric. To circumvent this problem, \cite{nokland2019training} computed the variance of the response over each channel in each convolutional layers, and then used it for cosine similarity.

We will use the same approach, which starts by arranging neurons into $c^k$ groups, labeled by $\alpha$, so that $z^k_{\alpha n}$ is the response of neuron $n$ in group $\alpha$ of layer $k$ (\cref{fig:circuitry}B).
We'll characterize each group by its ``smoothed'' variance (meaning we add a positive offset $\delta$), denoted $u_\alpha^k$,
\begin{align}
\label{smoothed_variance}
    u_\alpha^k \equiv \frac{\delta}{c^k_\alpha} +
    \frac{1}{c_\alpha^k} \sum_{n'} \brackets{ \boldring z_{\alpha n'}^k}^2;\quad \boldring z_{\alpha n}^k\equiv  z_{\alpha n}^k - \frac{1}{c_\alpha^k} \sum_{n'} z_{\alpha n'}^k,
\end{align}
where $c_\alpha^k$ is the number of neurons in group $\alpha$ of layer $k$.
One possible kernel would compare the standard deviation (so the squared root of $u_\alpha^k$) across different data points. However, we can get better performance by exponentiation and centering across channels. We thus define a new variable $v_\alpha^k$, 
\begin{align}
    v_\alpha^k \,&= (u_\alpha^k)^{1-p} - \frac{1}{c^k} \sum_{\alpha'} (u_{\alpha'}^k)^{1-p}
    \, ,
    \label{eq:grouping_v}
\end{align}
and use this grouped activity in the Gaussian kernel,
\begin{align}
\label{grouped_kernel}
    k(\bb z_i^k, \bb z_j^k) \,&= \exp\brackets{-\frac{1}{2\sigma^2}\norm{{\bb v}^k_i - {\bb v}^k_j}^2},
\end{align}
where, recall, $i$ and $j$ refer to different data points, and $(\bb v^k)_\alpha = v^k_\alpha\,$.

To illustrate the approach, we consider a linear network; see \cref{app:sec:update_derivations} for the general case.
Taking the derivative of $k(\bb z_i^k, \bb z_j^k)$ with respect to the weight in layer $k$, we arrive at
\begin{equation}
        \deriv{k(\bb z_i^k, \bb z_j^k)}{W^k_{\alpha n m}} \propto -\frac{1}{\sigma^2} k(\bb z_i^k, \bb z_j^k)\,({v}_{\alpha,\,i}^k - {v}_{\alpha,\,j}^k) \brackets{\frac{\boldring z^k_{\alpha n,\,j}}{(u_{\alpha,\,i}^k)^p} z^{k-1}_{m,\,i} - \frac{\boldring z^k_{\alpha n,\,j}}{(u_{\alpha,\,j}^k)^p} z^{k-1}_{m,\,j}}.
\end{equation}

Because of the term $u_\alpha^k$ in this expression, the learning rule is no longer strictly Hebbian. We can, though, make it Hebbian by assuming that the activity of the presynaptic neurons is $\boldring z^k_{\alpha n,\,j}/(u_{\alpha,\,j}^k)^p$,
\begin{equation}
    z^k_{\alpha n} = f\brackets{\sum_m W^k_{\alpha n m} r^{k-1}_m} ;\quad r^k_{\alpha n} \,= \frac{\boldring z_{\alpha n}^k}{(u_\alpha^k)^p}\,.
    \label{eq:divnorm_net}
\end{equation}
This automatically introduces divisive normalization into the network (\cref{fig:circuitry}C), a common ``canonical computation'' in the brain \cite{carandini2012normalization}, and in our case makes the update 3-factor Hebbian. It also changes the network from \cref{eq:feedforward_net} to \cref{eq:divnorm_net}, but that turns out to improve performance. The resulting update rule (again for a linear network; see \cref{app:subsec:gauss_divnorm_derivation} for the general case) is
\begin{equation}
\begin{split}
    \Delta W^k_{\alpha n m}  \,&\propto M^k_{\alpha,\,ij}\,\brackets{r^{k}_{\alpha n,\,i} r^{k-1}_{m,\,i} - r^{k}_{\alpha n,\,j} r^{k-1}_{m,\,j}}\,\\ M^k_{\alpha,\,ij}\,&=-\frac{1}{\sigma^2}\brackets{\gamma\, \boldring k(\bb y_i, \bb y_{j}) - 2\,\boldring k(\bb z^k_i,\bb z^k_j)} k(\bb z^k_i,\bb z^k_j)\,({v}_{\alpha,\,i}^k - {v}_{\alpha,\,j}^k)\, .
    \label{eq:divnorm_update}
\end{split}
\end{equation}
The circuitry to implement divisive normalization would be recurrent, but is out of scope of this work (see, however, \cite{grabska2017probabilistic} for network models of divisive normalization). For a convolutional layer, $\alpha$ would denote channels (or groups of channels), and the weights would be summed within each channel for weight sharing. When $p$, the exponent in \cref{eq:grouping_v}, is equal to 0.5, our normalization scheme is equivalent to divisive normalization in \cite{ren2016normalizing} and to group normalization \cite{wu2018group}.

Note that the rule is slightly different from the one for the basic Gaussian kernel (\cref{eq:gauss_update_ij}): the weight change is no longer proportional to differences of pre- and post-synaptic activities; instead, it is proportional to the difference in their product times the global factor for the channel. This form bears a superficial resemblance to Contrastive Hebbian learning \cite{movellan1991contrastive, xie2003equivalence}; however, that method doesn't have a third factor, and it generates points $i$ and $j$ using backprop-like feedback connections.

\subsection{Online update rules for the Gaussian kernel are standard Hebbian updates}
\label{sec:online_updates}
Our objective and update rules so far have used batches of data. However, we introduced pHSIC in \cref{sec:method} because a realistic network has to process data point by point. To show how this can work with our update rules, we first switch indices of points $i,\ j$ to time points $t,\ t-\Delta t$.

\subsubsection*{Gaussian kernel}

The update in \cref{eq:gauss_update_ij} becomes
\begin{equation}
    \Delta \bb W^k(t)  \propto M^k_{t,\,t-\Delta t} (\bb z^{k}_t - \bb z^{k}_{t-\Delta t}) (\bb z^{k-1}_t - \bb z^{k-1}_{t-\Delta t})\trans\,.
\label{eq:gauss_update_t}
\end{equation}
As an aside, if the point $t-\Delta t$ was presented for some period of time, its activity can be replaced by the mean activity: $\bb z^{k}_{t-\Delta t} \approx \pmb \mu^k_t$ and $z^{k-1}_{t-\Delta t} \approx \pmb \mu^{k-1}_t$. The update becomes a standard Hebbian one,
\begin{equation}
    \Delta \bb W^k(t)  \propto M^k_{t,\,t-\Delta t} (\bb z^{k}_t - \pmb\mu^k_t) (\bb z^{k-1}_t - \bb {\pmb\mu}^{k-1}_t)\trans\,.
\label{eq:gauss_update_t_hebb}
\end{equation}

\subsubsection*{Gaussian kernel with divisive normalization}
The update in \cref{eq:divnorm_update} allows two interpretations. The first one works just like before: we first introduce time, and then assume that the previous point $r^{k}_{\alpha n,\,t-\Delta t}$ is close to the short-term average of activity $\mu^k_{\alpha n,\,t}$. This results in
\begin{equation}
    \Delta W^k_{\alpha n m}(t)  \propto M^k_{\alpha,\,t\,,\,t-\Delta t}\,\brackets{r^{k}_{\alpha n,\,t} r^{k-1}_{m,\,t} - \mu^{k}_{\alpha n,\,t}\, \mu^{k-1}_{m,\,t}}\,.
    \label{eq:divnorm_update_t_hebb_first}
\end{equation}

The second one uses the fact that for points at times $t-\Delta t,\ t,\ t+\Delta t$ the Hebbian term of point $t$ appears twice: first as $M^k_{\alpha,\,t\,,\,t-\Delta t}\,r^{k}_{\alpha n,\,t} r^{k-1}_{m,\,t}$, and then as $-M^k_{\alpha,\,t+\Delta t\,,\,t}\,r^{k}_{\alpha n,\,t} r^{k-1}_{m,\,t}$. Therefore we can separate the Hebbian term at time $t$ and write the update as
\begin{equation}
    \Delta W^k_{\alpha n m}(t)  \propto \brackets{M^k_{\alpha,\,t\,,\,t-\Delta t} - M^k_{\alpha,\,t+\Delta t\,,\,t}}\,r^{k}_{\alpha n,\,t} r^{k-1}_{m,\,t}\,.
    \label{eq:divnorm_update_t_hebb_second}
\end{equation}

While the Hebbian part of \cref{eq:divnorm_update_t_hebb_second} is easier than in \cref{eq:divnorm_update_t_hebb_first}, it requires the third factor to span a longer time period. In both cases (and also for the plain Gaussian kernel), computing the third factor with local circuitry is relatively straightforward; for details see \cref{app:sec:synaptic_details}.

\section{Experiments}
\subsection{Experimental setup}
We compared our learning rule against stochastic gradient descent (SGD), and with an adaptive optimizer and batch normalization.  While networks with adaptive optimizers (e.g. Adam \cite{kingma2014adam}) and batch normalization (or batchnorm, \cite{ioffe2015batch}) perform better on deep learning tasks and are often used for biologically plausible algorithms (e.g. \cite{akrout2019deep,nokland2019training}), these features imply non-trivial ciruitry (e.g. the need for gradient steps in batchnorm). As our method focuses on what circuitry implements learning, the results on SGD match this focus better. 

We used the batch version of the update rule (\cref{eq:generic_update_rule}) only to make large-scale simulations computationally feasible. We considered the Gaussian kernel, and also the cosine similarity kernel, the latter to compare with previous work \cite{nokland2019training}. (Note, however, that the cosine similarity kernel gives implausible update rules, see \cref{app:sec:update_derivations}.) For both kernels we tested 2 variants of the rules: plain (without grouping), and with grouping and divisive normalization (as in \cref{eq:divnorm_update}). (Grouping \textit{without} divisive normalization performed as well or worse, and we don't report it here as the resulting update rules are less plausible; see \cref{app:sec:experiments}.) We also tested backprop, and learning of only the output layer in small-scale experiments to have a baseline result (also with divisive normalization in the network). In large-scale experiments, we also compared our approach to feedback alignment \cite{lillicrap2016random}, sign symmetry \cite{liao2015important} and layer-wise classification \cite{mostafa2018deep,nokland2019training}.

The nonlinearity was leaky ReLU (LReLU \cite{maas2013rectifier}; with a slope of $0.1$ for the negative input), but for convolutional networks trained with SGD we changed it to SELU \cite{klambauer2017self} as it performed better. All parameters (learning rates, learning rate schedules, grouping and kernel parameters) were tuned on a validation set (10\% of the training set). Optimizing HSIC instead of our approximation, pHSIC, didn't improve performance, and the original formulation of the HSIC bottleneck (\cref{eq:ma_objective}) performed much worse (not shown). The datasets were MNIST \cite{lecun2010mnist}, fashion-MNIST \cite{xiao2017fmnist}, Kuzushiji-MNIST \cite{clanuwat2018kmnist} and CIFAR10 \cite{krizhevsky2009cifar}. We used standard data augmentation for CIFAR10 \cite{krizhevsky2009cifar}, but no augmentation for the other datasets. All simulation parameters are provided in \cref{app:sec:experiments}. The implementation is available on GitHub: \href{https://github.com/romanpogodin/plausible-kernelized-bottleneck}{https://github.com/romanpogodin/plausible-kernelized-bottleneck}. 

\subsection{Small fully connected network}

We start with a 3-layer fully connected network (1024 neurons in each layer). To determine candidates for large-scale experiments, we compare the proposed rules to each other and to backprop. We thus delay comparison with other plausible learning methods to the next section; for performance of plausible methods in shallow networks see e.g. \cite{illing2019biologically}. The models were trained with SGD for 100 epochs with dropout \cite{srivastava2014dropout} of $0.05$, batch size of 256, and the bottleneck balance parameter $\gamma=2$ (other values of $\gamma$ performed worse); other parameters are provided in \cref{app:sec:experiments}.

Our results (summarized in \cref{tab:results:mlp} for mean test accuracy; see \cref{app:sec:experiments} for deviations between max and min) show a few clear trends across all four datasets: the kernel-based approaches with grouping and divisive normalization perform similarly to backprop on easy datasets (MNIST and its slightly harder analogues), but not on CIFAR10; grouping and divisive normalization had little effect on the cosine similarity performance; the Gaussian kernel required grouping and divisive normalization for decent accuracy. 
However, we'll see that the poor performance on CIFAR10 is not fundamental to the method; it's because the network is too small.

\subsection{Large convolutional networks and CIFAR10}
\begin{figure}[h]
  \centering
  \includegraphics[width=\textwidth]{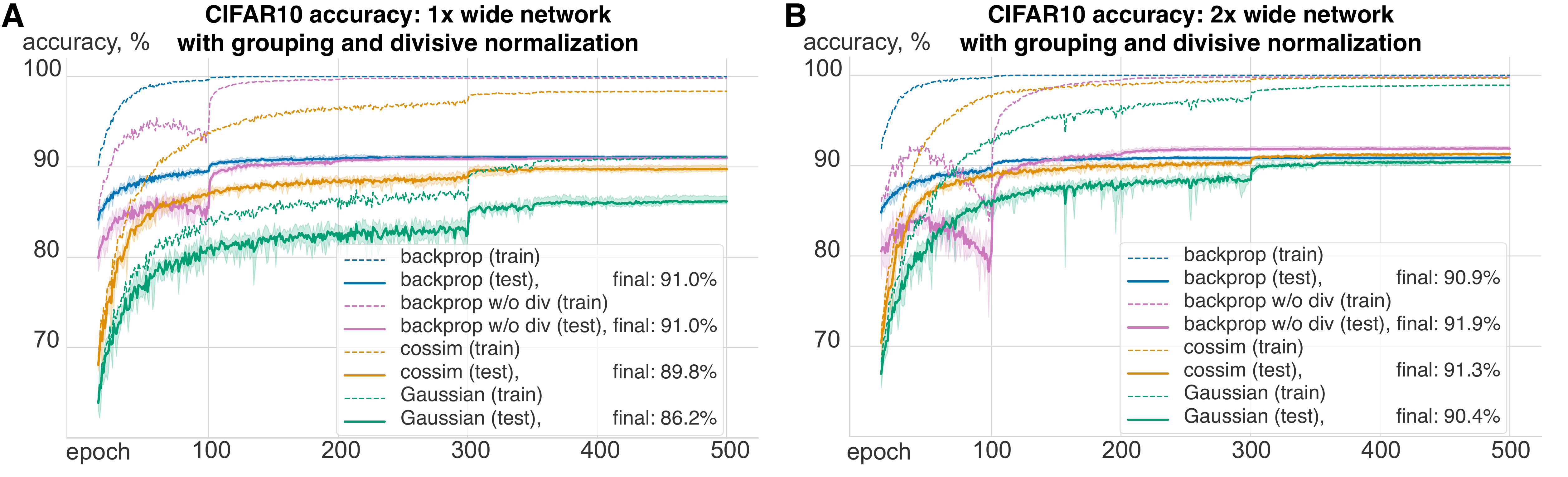}
  \caption{Performance of backprop, cosine similarity kernel (cossim) and Gaussian kernel on CIFAR10 with grouping and divisive normalization (and without for backprop; in pink). Solid lines: mean test accuracy over 5 random seeds; shaded areas: min/max test accuracy over 5 seeds; dashed lines: mean training accuracy over 5 seeds. \textbf{A.} 1x wide network: cosine similarity nearly matches backprop but doesn't achieve perfect training accuracy; Gaussian kernel lags behind cosine similarity. \textbf{B.} 2x wide network: backprop performance slightly improves; both kernels nearly match backprop performance, but Gaussian kernel still doesn't achieve perfect training accuracy.}
  \label{fig:results:vgg}
\end{figure}

Because all learning rules perform reasonably well on MNIST and its related extensions, in what follows we consider only CIFAR10, with the architecture used in \cite{nokland2019training}: \textbf{conv128-256-maxpool-256-512-maxpool-512-maxpool-512-maxpool-fc1024} and its version with double the convolutional channels (denoted 2x; the size of the fully connected layer remained the same). All networks were trained for 500 epochs with $\gamma=2$, dropout of $0.05$ and batch size of 128 (accuracy jumps in \cref{fig:results:vgg} indicate learning rate decreases); the rest of the parameters are provided in \cref{app:sec:experiments}.
\begin{table}[h]
  \caption{Mean test accuracy over 5 runs for a 3-layer (1024 neurons each) fully connected net. Last layer: training of the last layer; cossim: cosine similarity; grp: grouping; div: divisive normalization. 
  }
  \label{tab:results:mlp}
  \centering
  \begin{tabular}{lcccccccc}
    \toprule
    & \multicolumn{2}{c}{backprop} & \multicolumn{2}{c}{last layer} & \multicolumn{2}{c}{pHSIC: cossim} & \multicolumn{2}{c}{pHSIC: Gaussian}\\ \cmidrule(r){2-3}\cmidrule(r){4-5}\cmidrule(r){6-7}\cmidrule(r){8-9}& & div & & div & & grp+div  & & grp+div \\ \midrule MNIST & 98.6& 98.4& 92.0& 95.4& 94.9& 96.3& 94.6& 98.1\\fashion-MNIST & 90.2& 90.8& 83.3& 85.7& 86.3& 88.1& 86.5& 88.8\\Kuzushiji-MNIST & 93.4& 93.5& 71.2& 78.2& 80.4& 87.2& 80.2& 91.1\\CIFAR10 & 60.0& 60.3& 39.2& 38.0& 51.1&  47.6& 41.4& 46.4\\
    \bottomrule
  \end{tabular}
\end{table}
\begin{table}[h]
  \caption{Mean test accuracy on CIFAR10 over 5 runs for the 7-layer conv nets (1x and 2x wide). FA: feedback alignment; sign sym.: sign symmetry; layer class.: layer-wise classification; cossim: cosine similarity; divnorm: divisive normalization; bn: batchnorm.}
  \label{tab:results:vgg}
  \centering
  \begin{tabular}{lccccccc}
    \toprule
    & backprop & FA & sign sym. & \multicolumn{2}{c}{layer class.} & \multicolumn{2}{c}{pHSIC + grouping}\\\cmidrule(r){5-6}\cmidrule(r){7-8} & & &  & & +FA & cossim & Gaussian \\ \midrule
    1x + SGD + divnorm & 91.0 & 80.4 & 89.5 & 90.5 & 81.0 & 89.8 & 86.2\\
    2x + SGD + divnorm & 90.9 & 80.6 & 91.3 & 91.3 & 81.2 & 91.3 & 90.4 \\
    1x + AdamW + bn    & 94.1 & 82.4 & 93.6 & 92.1 & 90.3 & 91.3 & 89.9 \\
    2x + AdamW + bn    & 94.3 & 81.6 & 93.9 & 92.1 & 91.1 & 91.9 & 91.0 \\
    \bottomrule
  \end{tabular}
\end{table}

For SGD with divisive normalization (and also grouping for pHSIC-based methods; first two rows in \cref{tab:results:vgg}), layer-wise classification, sign symmetry and the cosine similarity kernel performed as well as backprop (consistent with previous findings \cite{nokland2019training, xiao2018biologically}); feedback alignment (and layer-wise classification with FA) performed significantly worse. In those cases, increasing the width of the network had a marginal effect on performance. The Gaussian kernel performed worse than backprop on the 1x network, but increasing the width closed the performance gap. A closer look at the learning dynamics of pHSIC-based methods and backprop (\cref{fig:results:vgg}) reveals low training accuracy with the Gaussian kernel on the 1x wide net, vs. almost 100\% for 2x, explaining low test accuracy. 

For AdamW \cite{loshchilov2017decoupled} with batchnorm \cite{ioffe2015batch} (last two rows in \cref{tab:results:vgg}), performance improved for all objectives, but it improved more for backprop (and sign symmetry) than for the pHSIC-based objectives. However, batch normalization (and in part adaptive learning rates of AdamW) introduces yet another implausible feature to the training method due to the batch-wide activity renormalization.

Only backprop, layer-wise classification (without feedback alignment) and the cosine similarity kernels performed well without any normalization (see \cref{app:sec:experiments}); for the other methods some kind of normalization was crucial for convergence. Backprop without divisive normalization (solid pink line in \cref{fig:results:vgg}) had non-monotonic performance, which can be fixed with a smaller learning rate at the cost of slightly worse performance.

\section{Discussion}
We proposed a layer-wise objective for training deep feedforward networks based on the kernelized information bottleneck, and showed that it can lead to biologically plausible 3-factor Hebbian learning. Our rules works nearly as well as backpropagation on a variety of image classification tasks. Unlike in classic Hebbian learning, where the pre- and post-synaptic activity triggers weight changes, our rules suggest large \textit{fluctuations} in this activity should trigger plasticity (e.g. when a new object appears).
 
Our learning rules do not need precise labels; instead they need only a binary signal: whether or not the previous and the current point have the same label. This allows networks to build representations with weaker supervision. We did train the last layer with precise labels, but that was only to compute accuracy; the network would learn just as well without it. To completely avoid supervision in hidden layers, it is possible to adapt our learning scheme to the contrastive (or self-supervised) setting as in Contrastive Predictive Coding \cite{oord2018representation, lowe2019putting} and SimCLR \cite{chen2020simple}.

Our rules do, though, need a global signal, which makes up part of the third factor. Where could it come from? In the case of the visual system, we can think of it as a ``teaching'' signal coming from other sensory areas. For instance, the more genetically pre-defined olfactory system might tell the brain that two successively presented object smell differently, and therefore should belong to different classes.
The third factor also contains a term that is local to the layer, but requires averaging of activity within the layer. This could be done by cells that influence plasticity, such as dopaminergic neurons \cite{gerstner2018eligibility} or glia \cite{todd2010perisynaptic}.

Although our approach is a step towards fully biologically plausible learning rules, it still suffers from some unrealistic features. The recurrence in our networks is limited to that necessary for divisive normalization, and has no excitatory within-layer recurrence or top-down signals. Those might be necessary to go from image classification to more realistic tasks (e.g., video). Our networks also allow negative firing rates due to the use of leaky ReLU and SELU nonlinearities. The latter (which we used to compensate for the lack of batchnorm) saturates for large negative inputs, and therefore the activity of each neuron can be viewed as a value relative to the background firing. Our main experiments also use convolutional networks, which are implausible due to weight sharing among neurons. Achieving good performance without weight sharing is an open question, although there are some results for backprop \cite{bartunov2018assessing}.


We showed that our rules can compete with backprop and its plausible approximations  on CIFAR10, even though they rely on less supervision and simpler error signals.
It should be possible to scale our learning rules to larger datasets, such as CIFAR100 \cite{krizhevsky2009cifar} and ImageNet \cite{deng2009imagenet}, as suggested by results from other layer-wise rules \cite{pmlr-v97-belilovsky19a,nokland2019training,lowe2019putting}. The layer-wise objectives can also make the theoretical analysis of deep learning easier. In fact, recent work analyzed a similar type of kernel-based objectives, showing its optimality with one hidden layer and backprop-comparable performance in deeper networks \cite{duan2020modularizing}.




The human brain contains about $10^{11}$ neurons, of which only about $10^6$ -- less than one per 100,000 -- are directly connected to the outside world; the rest make up hidden layers.
Understanding how such a system updates its synaptic strengths is one of the most challenging problems in neuroscience.
We proposed a family of biologically plausible learning rules for feedforward networks that have the potential to solve this problem. 
For a complete understanding of learning they need, of course, to be adapted to unsupervised and recurrent settings, and verified experimentally.
In addition, our learning rules are much more suitable for neuromorphic chips than standard backprop, due to the distributed nature of weight updates, and so could massively improve their scalability.

\section*{Broader Impact}
This research program, like most in neuroscience, has the potential to advance our understanding of the brain.
This comes with a host of societal implications. On the upside, it can give us a deeper understanding of mental health, possibly providing new therapies -- something that would improve the lives of on the order of 1 billion people. On the downside, a deeper understanding of the brain is likely to translate into accelerated development of artificial intelligence, which would put a great deal of power into the hands of a small number of people.

\section*{Acknowledgments}
This  work  was  supported  by  the  Gatsby  Charitable Foundation and the Wellcome Trust.

\bibliographystyle{unsrt}
\bibliography{bibliography}


\appendix
\renewcommand*\appendixpagename{\Large Appendices}
\appendixpage
\addappheadtotoc





\section{Kernel methods, HSIC and pHSIC}
\label{app:sec:kernels}
\subsection{Kernels and HSIC}
A kernel is a symmetric function $k(\bb a_1, \bb a_2)$ that maps $\RR^n\times \RR^n \rightarrow \RR$ and is positive-definite, 
\begin{equation}
    \forall\, \bb a_i\in \RR^n,\, \forall c_i\in\RR,\quad \sum_{ij}c_ic_jk(\bb a_i, \bb a_j) \geq 0\,.
\end{equation}
Consequently, the matrix $K_{ij}=k(\bb a_i,\bb a_j)$ is also positive-semidefinite.



We will use the following kernels,
\begin{alignat}{2}
    &\textrm{linear:}\qquad\qquad\qquad && k(\bb a_i,\bb a_j) = \bb a_i\trans\bb a_j\,;\\
    &\textrm{cosine similarity:} && k(\bb a_i,\bb a_j) = \bb a_i\trans\bb a_j / (\norm{\bb a_i}_2\norm{\bb a_j}_2)\,;\label{app:eq:cossim_kernel}\\
    &\textrm{Gaussian:} && k(\bb a_i,\bb a_j) = \exp(-\norm{\bb a_i-\bb a_j}^2_2 / (2\sigma^2))\,.\label{app:eq:gaussian_kernel}
\end{alignat}

The Hilbert-Schmidt Independence Criterion (introduced in \cite{gretton2005measuring}) measures independence between two random variable, $A$ and $B$, with kernels $k_a$ and $k_b$,
\begin{align}
    \hsic{A}{B} \,&= \brackets{\expect_{\bb a_1\bb b_1}\expect_{\bb a_2\bb b_2} - 2\,\expect_{\bb a_1\bb b_1}\expect_{\bb a_2}\expect_{\bb b_2} + \expect_{\bb a_1}\expect_{\bb b_1}\expect_{\bb a_2}\expect_{\bb b_2}}\brackets{k(\bb a_1, \bb a_2) \,k(\bb b_1,\bb b_2)},
    \label{app:eq:hsic2}
\end{align}
where $\expect_{\bb a_1\bb b_1}$ denotes expectation over the joint distribution.

When $B\equiv A$, meaning that $p_{ab}(\bb a,\bb b)=p_{a}(\bb a)\,\delta(\bb b-\bb a)$, HSIC becomes
\begin{align}
    \hsic{A}{A} \,&=\expect_{\bb a_1}\expect_{\bb a_2}\brackets{k(\bb a_1, \bb a_2)}^2 - 2\,\expect_{\bb a_1}\brackets{\expect_{\bb a_2}k(\bb a_1, \bb a_2)}^2 + \brackets{\expect_{\bb a_1}\expect_{\bb a_2}k(\bb a_1, \bb a_2)}^2\label{app:eq:hsic_aa_expect}\,.
\end{align}

If both kernels are linear, it is easy to show that HSIC becomes the squared Frobenius norm of the cross-covariance,
\begin{align}
    \hsic{A}{B} \,&= \norm{\bb C_{\bb a\bb b}}_F^2,\quad \bb C_{\bb a\bb b} = \expect_{\bb a\bb b}\,\bb a \bb b\trans\,.
\end{align}


In general, HSIC follows the same intuition -- it is the squared Hilbert-Schmidt norm (generalization of the Frobenius norm) of the cross-covariance operator.

\subsection{pHSIC}
\label{app:subsec:phsic}
We define the ``plausible'' HSIC by substituting $\expect_{\bb a_1\bb b_1}\expect_{\bb a_2}\expect_{\bb b_2}$ in \cref{app:eq:hsic2} by $\expect_{\bb a_1}\expect_{\bb b_1}\expect_{\bb a_2}\expect_{\bb b_2}$,
\begin{align}
    \mathrm{p}\hsic{A}{B} \,&=    \brackets{\expect_{\bb a_1\bb b_1}\expect_{\bb a_2\bb b_2} - \expect_{\bb a_1}\expect_{\bb b_1}\expect_{\bb a_2}\expect_{\bb b_2}}\brackets{k(\bb a_1, \bb a_2) \,k(\bb b_1,\bb b_2)}
    \, .
    \label{app:eq:hsic_plausible}
\end{align}
Therefore, $\hsic{A}{B}=\mathrm{p}\hsic{A}{B}$ when $\bb a$ and $\bb b$ are independent ($\expect_{\bb a_1\bb b_1}=\expect_{\bb a_1}\expect_{\bb b_1}$, which is not useful in our case), and when $\expect_{\bb a_2}k(\bb a_1,\bb a_2)=0$ or $\expect_{\bb b_2}k(\bb b_1,\bb b_2)=0$. 

In addition, $\mathrm{p}\hsic{A}{A}$ becomes the variance of $k(\bb a_1, \bb a_2)$ with respect to $p_a(\bb a_1)p_a(\bb a_2)$, 
\begin{align}
\mathrm{p}\hsic{A}{A} =
\expect_{\bb a_1}\expect_{\bb a_2} \brackets{k(\bb a_1, \bb a_2) }^2 -
\brackets{
\expect_{\bb a_1}\expect_{\bb a_2} k(\bb a_1, \bb a_2) }^2=\VV{k(\bb a_1, \bb a_2)}\, .
\label{app:eq:phsic_aa}
\end{align}
By combining \cref{app:eq:hsic_aa_expect} and \cref{app:eq:phsic_aa}, we can show that $\hsic{A}{A}\leq\mathrm{p}\hsic{A}{A}$: denoting $\mu_a(\bb a_1)=\expect_{\bb a_2}k(\bb a_1,\bb a_2)$, we have
\begin{align}
\begin{split}
    \mathrm{p}\hsic{A}{A} - \hsic{A}{A} \,& = 2\,\expect_{\bb a_1}\brackets{\expect_{\bb a_2}k(\bb a_1, \bb a_2)}^2 - 2\,\brackets{\expect_{\bb a_1}\expect_{\bb a_2} k(\bb a_1, \bb a_2) }^2\\
    &=2\brackets{\expect_{\bb a_1}\mu_a(\bb a_1)^2 - \brackets{\expect_{\bb a_1}\mu_a(\bb a_1) }^2} \\
    &=2\,\VV{\mu_a(\bb a_1)} \geq 0.
\end{split}
\end{align}

As a result, our objective,
\begin{equation}
    \mathrm{p}\hsic{Z^k}{Z^k} - \gamma\, \mathrm{p}\hsic{Y}{Z^k}\,,
\end{equation}
is an upper bound on the ``true'' objective whenever $\mathrm{p}\hsic{Y}{Z^k} =\hsic{Y}{Z^k}$, which is the case when $\expect_{\bb y_2}k(\bb y_1,\bb y_2)=0$.

Finally, the empirical estimate of pHSIC (which can be derived just like for HSIC in \cite{gretton2005measuring}) is
\begin{align}
    \mathrm{p}\widehat{\mathrm{HSIC}}(A,B) \,& =\frac{1}{m^2}\sum_{ij} k(\bb a_i, \bb a_j) k(\bb b_i, \bb b_j)
    -
    \frac{1}{m^2}\sum_{ij} k(\bb a_i, \bb a_j)
    \frac{1}{m^2}\sum_{ql} k(\bb b_q, \bb b_l)\,.
    \label{app:eq:phsic_bar}
\end{align}

\subsection{How much information about the label do we need?}
\label{app:subsec:binary_teaching}

In our rules, the information about the label comes only through $k(\bb y_i,\bb y_j)$, where $\bb y$ is a one-hot vector -- for $n$ classes, an $n$-dimensional vector of mainly zeros with only a single one (which corresponds to its label). 
For $k(\bb y_i,\bb y_j)$ we use the cosine similarity kernel, with $\bb y$ centered. If the dataset is balanced (i.e., all classes have the same probability, $1/n$), $\expect_{\bb y_j}\,k(\bb y_i,\bb y_j)=0$ and the resulting kernel is
\begin{align}
    k(\bb y_i,\bb y_j) = \frac{(\bb y_i - \frac{1}{n}\bb 1_n)\trans(\bb y_j - \frac{1}{n}\bb 1_n)}{\norm{\bb y_i - \frac{1}{n}\bb 1_n}\norm{\bb y_j - \frac{1}{n}\bb 1_n}}  = \frac{\II{\bb y_i = \bb y_j} - \frac{1}{n}}{1 - \frac{1}{n}} = \begin{cases}
        1,& \bb y_i=\bb y_j\,,\\
        -\frac{1}{n-1},& \mathrm{otherwise}\,.
    \end{cases}
    \label{eq:y_kernel_balanced}
\end{align}
If there are many classes, this signal approaches $\II{\bb y_i =\bb y_j}$, which is what the same as the uncentered linear kernel.

Equation~\eqref{eq:y_kernel_balanced} is especially convenient, because the kernel takes on only two values, $1$ and $-1/(n-1)$. Consequently, precise labels are not needed.
This is not the case for unbalanced classes, as $\expect \bb y_i \neq \bb 1_n / n$ and centering of $\bb y$ doesn't make the normalized vector centered.  However, taking the linear kernel gives an almost binary signal,
\begin{equation}
\begin{split}
    k(\bb y_i,\bb y_j) \,&= (\bb y_i - \bb p)\trans(\bb y_j - \bb p) = \II{\bb y_i =\bb y_j} + \sum_k p_k^2 - p_i - p_j \\
    &= \II{\bb y_i =\bb y_j} + \bigO{\frac{1}{n}},
\end{split}
\end{equation}
as long as the probability of each class $k$, $p_k$, is $\bigO{1/n}$ (i.e., they are roughly balanced). As a result, the teaching signal is nearly binary, and we can compute it without knowing the probability of each class.

\section{Derivations of the update rules for plausible kernelized information bottleneck}
\label{app:sec:update_derivations}
\subsection{General update rule}
Here we derive the gradient of pHSIC in our network, along with its empirical estimate.
Our starting point is the observation that because the network is feedforward, the activity in layer $k$ is a deterministic function of the previous layer and the weights: $Z^k = f\brackets{\bb W^k, Z^{k-1}}$ (this includes both feedforward layers and layers with divisive normalization). Therefore, we can write the expectations of any function $g(Y, Z^k)$ (which need not actually depend on $Y$) in terms of $Z^{k-1}$,
\begin{align}
    \expect_{\bb y\bb z^k} g(\bb y,\bb z^k) \,
    = \expect_{\bb y\bb z^{k-1}} g\brackets{\bb y, f\brackets{\bb W^k, \bb z^{k-1}}} \, .
\end{align}
As a result, we can write the gradients of pHSIC (assuming we can exchange the order of differentiation and expectation, and the function is differentiable at $\bb W^k$) as
\begin{subequations}
\begin{align}
\begin{split}
    & \, \, \, \, \deriv{\mathrm{p}\hsic{Y}{Z^k}}{\bb W^k} =
    \brackets{\expect_{\bb y_1\bb z_1^{k-1}}\expect_{\bb y_2\bb z_2^{k-1}} -\expect_{\bb y_1}\expect_{\bb z_1^{k-1}}\expect_{\bb y_2}\expect_{\bb z_2^{k-1}}}
\\
    & \quad \quad \quad \quad \quad \quad \quad \quad \quad \quad  
    k(\bb y_1, \bb y_2) \,
    \deriv{k\brackets{ f\brackets{\bb W^k, \bb z^{k-1}_1},f\brackets{\bb W^k, \bb z^{k-1}_2}}}{\bb W^k}\,,
\end{split}
\\
\begin{split}
    & \deriv{\mathrm{p}\hsic{Z^k}{Z^k}}{\bb W^k} =
    2\,\expect_{\bb z_1^{k-1}}\expect_{\bb z_2^{k-1}} k(\bb z_1^{k}, \bb z_2^{k})
    \deriv{k\brackets{ f\brackets{\bb W^k, \bb z^{k-1}_1},f\brackets{\bb W^k, \bb z^{k-1}_2}}}{\bb W^k}
 \\
    & \quad \quad \quad \quad  
    -2 \,
    \brackets{\expect_{\bb z_1^{k-1}} \expect_{\bb z_2^{k-1}}k(\bb z_1^{k}, \bb z_2^{k-1})} \brackets{\expect_{\bb z_1^{k-1}}\expect_{\bb z_2^{k-1}}\deriv{k\brackets{ f\brackets{\bb W^k, \bb z^{k-1}_1},f\brackets{\bb W^k, \bb z^{k-1}_2}}}{\bb W^k} }
    \, .
\end{split}
\end{align}
\end{subequations}

To compute these derivates from data, we take empirical averages (see \cref{app:eq:phsic_bar}),
\begin{equation}
\begin{split}
    &\deriv{\brackets{\mathrm{p}\widehat{\mathrm{HSIC}}(Z^k,Z^k) - \gamma\, \mathrm{p}\widehat{\mathrm{HSIC}}(Y,Z^k)}}{\bb W^k}  =\\
    &\qquad2\,\frac{1}{m^2}\sum_{ij} k(\bb z_i^k, \bb z_j^k) \deriv{k(\bb z_i^k, \bb z_j^k)}{\bb W^k}
    -
    2\,\frac{1}{m^2}\sum_{ql} k(\bb z_q^k, \bb z_l^k)
    \frac{1}{m^2}\sum_{ij} \deriv{k(\bb z_i^k, \bb z_j^k)}{\bb W^k}\\
    &\qquad-\gamma\,\frac{1}{m^2}\sum_{ij} k(\bb y_i, \bb y_j) \deriv{k(\bb z_i^k, \bb z_j^k)}{\bb W^k}
    +\gamma\,
    \frac{1}{m^2}\sum_{ql} k(\bb y_q, \bb y_l)
    \frac{1}{m^2}\sum_{ij} \deriv{k(\bb z_i^k, \bb z_j^k)}{\bb W^k}.
\end{split}
\end{equation}
Making the definition $\boldring k(\bb a_i, \bb a_j) = k(\bb a_i, \bb a_j) - \sum_{ql}k(\bb a_q, \bb a_l) / m^2$, this simplifies to
\begin{equation}
\begin{split}
    &\deriv{\brackets{\mathrm{p}\widehat{\mathrm{HSIC}}(Z^k,Z^k) - \gamma\, \mathrm{p}\widehat{\mathrm{HSIC}}(Y,Z^k)}}{\bb W^k} \\
    &\qquad\qquad= \frac{1}{m^2} \sum_{ij}\brackets{2\,\boldring k(\bb z_i^k, \bb z_j^k) - \gamma\,\boldring k(\bb y_i, \bb y_j)}\deriv{k(\bb z_i^k, \bb z_j^k)}{\bb W^k}\,.
    \label{app:eq:phsic_obj_deriv}
\end{split}
\end{equation}

The key quantity in the above expressions is the derivative of the kernel with respect to the weights. Below we compute those for the Gaussian and cosine similarity kernels, and explain why the cosine similarity update is implausible.


\subsection{Gaussian kernel}
\label{app:subsec:gauss_derivation}

For the Gaussian kernel, the derivative with respect to a single weight is
\begin{align}
\begin{split}
    \deriv{k(\bb z_i^k, \bb z_j^k)}{W^k_{nm}} \,& = \deriv{}{ W^k_{nm}}\exp\brackets{-\frac{1}{2\sigma ^2}\norm{\bb z_i^k - \bb z_j^k}^2} \\
    &= -\frac{k(\bb z_i^k, \bb z_j^k)}{\sigma^2} \brackets{z_{n,\,i}^k - z_{n,\,j}^k}\deriv{\brackets{z_{n,\,i}^k - z_{n,\,j}^k}}{ W^k_{nm}}\\
    &=-\frac{k(\bb z_i^k, \bb z_j^k)}{\sigma^2} \brackets{z_{n,\,i}^k - z_{n,\,j}^k}\brackets{f'\brackets{\bb W^k \bb z^{k-1}_i}_n z^{k-1}_{m,\,i} - f'\brackets{\bb W^k \bb z^{k-1}_j}_n z^{k-1}_{m,\,j}}.
\end{split}
\end{align}
For the linear network $f'(x)=1$, and so the gradient w.r.t. $\bb W^k$ becomes an outer product (\cref{eq:gauss_update_ij}).

\subsection{Gaussian kernel with grouping and divisive normalization}
\label{app:subsec:gauss_divnorm_derivation}
For the circuit with grouping and divisive normalization, the kernel is a function of $\bb v$, not $\bb z$ (see Eqs.~\eqref{smoothed_variance}, \eqref{eq:grouping_v} and \eqref{grouped_kernel}). This makes the derivative with respect to $\bb z$ more complicated than the above expression would suggest. Specifically, using \cref{grouped_kernel} for the kernel with grouping, we have
\begin{align}
\begin{split}
    \deriv{k(\bb z_i^k, \bb z_j^k)}{W^k_{\alpha n m}} \,&=\deriv{}{W^k_{\alpha n m}}\exp\brackets{-\frac{1}{2\sigma^2}\norm{{\bb v}^k_i - {\bb v}^k_j}^2}\\
    &=-\frac{k(\bb z_i^k, \bb z_j^k)}{\sigma^2} \sum_{\alpha '}({v}_{\alpha',i}^k - {v}_{\alpha',\,j}^k)\deriv{({v}_{\alpha',\,i}^k - {v}_{\alpha',\,j}^k)}{W^k_{\alpha n m}}\,,
    \label{app:eq:grouping_derivx}
\end{split}
\end{align}
where the sum over $\alpha'$ appears due to centering (so all $\alpha$ are coupled). Using \cref{eq:grouping_v} to express $\bb v$ in terms of $\bb u$, the above derivative is
\begin{align}
    \deriv{{v}_{\alpha',\,i}^k}{W^k_{\alpha n m}}\,
    & =
    \deriv{\brackets{(u_{\alpha',\,i}^k)^{1-p} - \frac{1}{c_\alpha^k} \sum_{\alpha''} (u_{\alpha'',\,i}^k)^{1-p}}}{W^k_{\alpha n m}} = \brackets{\delta_{\alpha\alpha'} - \frac{1}{c^k_\alpha}}\deriv{(u_{\alpha,\,i}^k)^{1-p}}{W^k_{\alpha n m}} \, .
    \label{app:eq:grouping_v_derivx}
\end{align}
Then using \cref{smoothed_variance} to express $\bb u$ in terms of $\bb z$, we have
\begin{align}
\begin{split}
    \deriv{(u_{\alpha,\,i}^k)^{1-p}}{W^k_{\alpha n m}} \,&=\frac{1-p}{(u_{\alpha,\,i}^k)^{p}}\deriv{\brackets{\frac{\delta}{c^k_\alpha} +
    \frac{1}{c_\alpha^k} \sum_{n'} \brackets{ \boldring z_{\alpha n',\,i}^k}^2}}{W^k_{\alpha n m}}
    \\
    &=\frac{1-p}{(u_{\alpha,\,i}^k)^{p}}\frac{2}{c_\alpha^k}\sum_{n'}\boldring z_{\alpha n',\,i}^k\deriv{\boldring z_{\alpha n',\,i}^k}{W^k_{\alpha n m}}\\ &=\frac{1-p}{(u_{\alpha,\,i}^k)^{p}}\frac{2}{c_\alpha^k}\sum_{n'}\boldring z_{\alpha n',\,i}^k\deriv{\brackets{z_{\alpha n',\,i}^k - \frac{1}{c_\alpha^k} \sum_{n''} z_{\alpha n'',\,i}^k}}{W^k_{\alpha n m}}\\
    &=\frac{1-p}{(u_{\alpha,\,i}^k)^{p}}\frac{2}{c_\alpha^k}\sum_{n'}\boldring z_{\alpha n',\,i}^k \brackets{\delta_{n'n} - \frac{1}{c_\alpha^k}} \deriv{z_{\alpha n,\,i}^k}{W^k_{\alpha n m}}
    \\
    &=\frac{1-p}{(u_{\alpha,\,i}^k)^{p}}\frac{2}{c_\alpha^k}\boldring z_{\alpha n,\,i}^k \deriv{z_{\alpha n,\,i}^k}{W^k_{\alpha n m}} \, .
\end{split}
\end{align}
Inserting this expression into \cref{app:eq:grouping_v_derivx}, inserting that into \cref{app:eq:grouping_derivx}, and performing a small amount of algebra, we arrive at
\begin{align}
    \deriv{k(\bb z_i^k, \bb z_j^k)}{W^k_{\alpha n m}} \,&=-\frac{k(\bb z_i^k, \bb z_j^k)}{\sigma^2} \sum_{\alpha '}({v}_{\alpha',\,i}^k - {v}_{\alpha',\,j}^k)\brackets{\delta_{\alpha\alpha'} - \frac{1}{c^k_\alpha}}\frac{2\,(1-p)}{c_\alpha^k}\\
    &\qquad\times\brackets{\frac{\boldring z_{\alpha n,\,i}^k}{(u_{\alpha,\,i}^k)^{p}} \deriv{z_{\alpha n,\,i}^k}{W^k_{\alpha n m}} - \frac{\boldring z_{\alpha n,\,j}^k}{(u_{\alpha,\,j}^k)^{p}} \deriv{z_{\alpha n,\,j}^k}{W^k_{\alpha n m}}}.
\end{align}
As $v^k_{\alpha,\,i}$ is centered with respect to $\alpha$ and the last term doesn't depend on $\alpha'$, the final expression becomes
\begin{align}
    \deriv{k(\bb z_i^k, \bb z_j^k)}{W^k_{\alpha n m}} \,&=-\frac{2\,(1-p)\,k(\bb z_i^k, \bb z_j^k)}{\sigma^2c_\alpha^k} ({v}_{\alpha,\,i}^k - {v}_{\alpha,\,j}^k)\\
    &\qquad\times\brackets{\frac{\boldring z_{\alpha n,\,i}^k}{(u_{\alpha,\,i}^k)^{p}} f'\brackets{\bb W^k \bb z^{k-1}_i}_n z^{k-1}_{m,\,i} - \frac{\boldring z_{\alpha n,\,j}^k}{(u_{\alpha,\,j}^k)^{p}} f'\brackets{\bb W^k \bb z^{k-1}_j}_n z^{k-1}_{m,\,j}}.
\end{align}

\subsection{Cosine similarity kernel}
\label{app:subsec:cossim_derivation}
Assuming that $\bb z^k$ is bounded away from zero (because $\bb z^k / \norm{\bb z^k}$ is not continuous at 0; adding a smoothing term to the norm would help but won't change the derivation), the derivative of the cosine similarity kernel is
\begin{align}
\begin{split}
    \deriv{k(\bb z_i^k, \bb z_j^k)}{W^k_{nm}} \,& = \deriv{}{W^k_{nm}}\frac{(\bb z_i^k)\trans \bb z_j^k}{\norm{\bb z_i^k}\norm{\bb z_j^k}} \\
    &= \frac{1}{\norm{\bb z_i^k}\norm{\bb z_j^k}}\deriv{\brackets{z_{n,\,i}^k z_{n,\,j}^k}}{W^k_{nm}} - \frac{(\bb z_i^k)\trans \bb z_j^k}{\norm{\bb z_i^k}^2\norm{\bb z_j^k}^2}\deriv{\norm{\bb z_i^k}\norm{\bb z_j^k}}{W^k_{nm}}\,.
\end{split}
\end{align}
The first derivative is simple,
\begin{align}
    \deriv{\brackets{z_{n,\,i}^k z_{n,\,j}^k}}{W^k_{nm}}\,&=z_{n,\,i}^k\,f'\brackets{\bb W^k \bb z^{k-1}_j}_n z^{k-1}_{m,\,j} + z_{n,\,j}^k\,f'\brackets{\bb W^k \bb z^{k-1}_i}_n z^{k-1}_{m,\,i}\,.
    \label{app:eq:cossim_deriv_first_term}
\end{align}
However, in this form it is hard to interpret, as both terms have the pre-synaptic activity at one point and the post-synaptic activity at the other. We can, though, can re-arrange it into differences in activity,
\begin{align}
\begin{split}
    \deriv{\brackets{z_{n,\,i}^k z_{n,\,j}^k}}{W^k_{nm}}\,&=\brackets{z_{n,\,i}^k - z_{n,\,j}^k}\,f'\brackets{\bb W^k \bb z^{k-1}_j}_n z^{k-1}_{m,\,j} + z_{n,\,j}^k\,f'\brackets{\bb W^k \bb z^{k-1}_j}_n z^{k-1}_{m,\,j}\\
    &\quad+ \brackets{z_{n,\,j}^k - z_{n,\,i}^k}\,f'\brackets{\bb W^k \bb z^{k-1}_i}_n z^{k-1}_{m,\,i} + z_{n,\,i}^k\,f'\brackets{\bb W^k \bb z^{k-1}_i}_n z^{k-1}_{m,i}\\
    &=-\brackets{z_{n,\,i}^k - z_{n,\,j}^k}\,\brackets{f'\brackets{\bb W^k \bb z^{k-1}_i}_n z^{k-1}_{m,\,i} - f'\brackets{\bb W^k \bb z^{k-1}_j}_n z^{k-1}_{m,\,j}}\\
    &\quad+z_{n,\,i}^k\,f'\brackets{\bb W^k \bb z^{k-1}_i}_n z^{k-1}_{m,\,i}+z_{n,\,j}^k\,f'\brackets{\bb W^k \bb z^{k-1}_j}_n z^{k-1}_{m,\,j}\,.
\end{split}
\end{align}

The second one is slightly harder,
\begin{align}
\begin{split}
    \deriv{\norm{\bb z_i^k}\norm{\bb z_j^k}}{W^k_{nm}}\,&= \norm{\bb z_i^k}\deriv{\sqrt{\sum_{n'} (z_{n',\,j}^k)^2}}{W^k_{nm}} + \norm{\bb z_j^k}\deriv{\sqrt{\sum_{n'} (z_{n',\,i}^k)^2}}{W^k_{nm}}\\
    &=\frac{\norm{\bb z_i^k}}{\norm{\bb z_j^k}}\,z^k_{n,\,j}\deriv{z^k_{n,\,j}}{W^k_{nm}} + \frac{\norm{\bb z_j^k}}{\norm{\bb z_i^k}}z^k_{n,\,i}\deriv{z^k_{n,\,i}}{W^k_{nm}}\\
    &=\frac{\norm{\bb z_i^k}}{\norm{\bb z_j^k}}\,z^k_{n,\,j}f'\brackets{\bb W^k \bb z^{k-1}_j}_n z^{k-1}_{m,\,j} + \frac{\norm{\bb z_j^k}}{\norm{\bb z_i^k}}z^k_{n,\,i}f'\brackets{\bb W^k \bb z^{k-1}_i}_n z^{k-1}_{m,\,i}\,.
\end{split}
\end{align}
Grouping these results together,
\begin{align}
\begin{split}
    \deriv{k(\bb z_i^k, \bb z_j^k)}{W^k_{nm}} \,&  =-\frac{\brackets{z_{n,\,i}^k - z_{n,\,j}^k}\,\brackets{f'\brackets{\bb W^k \bb z^{k-1}_i}_n z^{k-1}_{m,\,i} - f'\brackets{\bb W^k \bb z^{k-1}_j}_n z^{k-1}_{m,\,j}}}{\norm{\bb z_i^k}\norm{\bb z_j^k}}\\
    &\quad+\frac{z_{n,\,i}^k\,f'\brackets{\bb W^k \bb z^{k-1}_i}_n z^{k-1}_{m,\,i}}{\norm{\bb z_i^k}\norm{\bb z_j^k}}+\frac{z_{n,\,j}^k\,f'\brackets{\bb W^k \bb z^{k-1}_j}_n z^{k-1}_{m,\,j}}{\norm{\bb z_i^k}\norm{\bb z_j^k}}\\
    &\quad- \frac{(\bb z_i^k)\trans \bb z_j^k}{\norm{\bb z_i^k}^2\norm{\bb z_j^k}^2}\frac{\norm{\bb z_i^k}}{\norm{\bb z_j^k}}\,z^k_{n,\,j}f'\brackets{\bb W^k \bb z^{k-1}_j}_n z^{k-1}_{m,\,j} \\
    &\quad- \frac{(\bb z_i^k)\trans \bb z_j^k}{\norm{\bb z_i^k}^2\norm{\bb z_j^k}^2}\frac{\norm{\bb z_j^k}}{\norm{\bb z_i^k}}\,z^k_{n,\,i}f'\brackets{\bb W^k \bb z^{k-1}_i}_n z^{k-1}_{m,\,i}\,.
\end{split}
\end{align}
As all terms share the same divisor, we can write the last expression more concisely as
\begin{align}
\begin{split}
    \norm{\bb z_i^k}\norm{\bb z_j^k}\deriv{k(\bb z_i^k, \bb z_j^k)}{W^k_{nm}} \,&  =-\brackets{z_{n,\,i}^k - z_{n,\,j}^k}\,\brackets{f'\brackets{\bb W^k \bb z^{k-1}_i}_n z^{k-1}_{m,\,i} - f'\brackets{\bb W^k \bb z^{k-1}_j}_n z^{k-1}_{m,\,j}}\\
    &\quad+\sum_{s=i,\,j}\brackets{1 - \frac{(\bb z_i^k)\trans \bb z_j^k}{\norm{\bb z_s^k}^2}}z_{n,\,s}^k\,f'\brackets{\bb W^k \bb z^{k-1}_s}_n z^{k-1}_{m,\,s}\,.
\end{split}
\end{align}

Considering a linear network for simplicity, the weight update over two points $i,\,j$ (negative of a single term in the sum in \cref{app:eq:phsic_obj_deriv}) for the cosine similarity kernel becomes 
\begin{align}
\begin{split}
    \Delta W^k_{nm}\,&= M_{ij}^k\brackets{z_{n,\,i}^k - z_{n,\,j}^k}\,\brackets{z^{k-1}_{m,\,i} - z^{k-1}_{m,\,j}} - M_{ij}^k\sum_{s=i,\,j}\brackets{1 - \frac{(\bb z_i^k)\trans \bb z_j^k}{\norm{\bb z_s^k}^2}}z_{n,\,s}^k z^{k-1}_{m,\,s},\label{app:eq:cossim_two_points}\\
    M_{ij}^k \,& = \frac{1}{\norm{\bb z_i^k}\norm{\bb z_j^k}} \brackets{2\,\boldring k(\bb z_i^k, \bb z_j^k) - \gamma\,\boldring k(\bb y_i, \bb y_j)}.
\end{split}
\end{align}

This rule is biologically implausible (or very hard to implement) for two reasons.
First, to compute $M_{ij}^k$ the layer needs to track three signal simultaneously: $(\bb z_i^k)\trans \bb z_j^k$, $\norm{\bb z_i^k}$ and $\norm{\bb z_j^k}$ (versus only one for the Gaussian kernel, $\norm{\bb z_i^k - \bb z_j^k}$).
Second, assuming point $i$ comes after $j$, the pre-factor of the Hebbian term (the sum in \cref{app:eq:cossim_two_points}) for $s=j$ can't be computed until the networks receives point $i$. As a result, the update requires three independent plasticity pathways (one for $M_{ij}^k\brackets{z_{n,\,i}^k - z_{n,\,j}^k}\,\brackets{z^{k-1}_{m,\,i} - z^{k-1}_{m,\,j}}$ and two for the sum in \cref{app:eq:cossim_two_points}), because each term in \cref{app:eq:cossim_two_points} combines the pre- and post-synaptic activity on different timescales and with different third factors.

Adding grouping and divisive normalization to this rule is straightforward: we need to use the grouped response $v_\alpha^k$ (\cref{eq:grouping_v}) in the kernel as $k(\bb z^k_i,\bb z^k_j) = (\bb v_i^k)\trans\bb v_j^k / (\norm{\bb v_i^k}\norm{\bb v_j^k})$, and repeat the calculation above (divisive normalization will appear here too as it results from differentiating $v_\alpha^k$). As it would only make the circuitry more complicated, we omit the derivation. Note that if we use grouping with $p=0.5$ but don't introduce divisive normalization, our objective resembles the ``sim-bpf'' loss in \cite{nokland2019training} (but it doesn't match it exactly, as we also introduce centering of the kernel and group convolutional layers over multiple channels rather than one).

\subsection{Linear kernel}
Derivation of the update for the linear kernel only needs the derivative from \cref{app:eq:cossim_deriv_first_term}, therefore by doing the same calculations we obtain
\begin{align}
\begin{split}
    \deriv{k(\bb z_i^k, \bb z_j^k)}{W^k_{nm}} \,& = \deriv{(\bb z_i^k)\trans \bb z_j^k}{W^k_{nm}} =\deriv{\brackets{z_{n,\,i}^k z_{n,\,j}^k}}{W^k_{nm}}\\
    &=-\brackets{z_{n,\,i}^k - z_{n,\,j}^k}\,\brackets{f'\brackets{\bb W^k \bb z^{k-1}_i}_n z^{k-1}_{m,\,i} - f'\brackets{\bb W^k \bb z^{k-1}_j}_n z^{k-1}_{m,\,j}}\\
    &\quad+\sum_{s=i,\,j}z_{n,\,s}^k\,f'\brackets{\bb W^k \bb z^{k-1}_s}_n z^{k-1}_{m,\,s}.
\end{split}
\end{align}
This is much easier to compute than the cosine similarity kernel: it only uses $(\bb z_i^k)\trans \bb z_j^k$ in the third factor, and needs two plasticity channels rather than three (as now both terms in the sum use the same third factor). However, we couldn't achieve good performance with this kernel.

\section{Circuitry to implement the update rules}
\label{app:sec:synaptic_details}
In this section we outline the circuitry needed to compute the Hebbian terms and the third factor for the Gaussian kernel (plain and with grouping and divisive normalization).

\subsection{Hebbian terms}
\label{app:subsec:gauss_synaptic_details}
\begin{figure}[h]
  \centering
  \includegraphics[width=\textwidth]{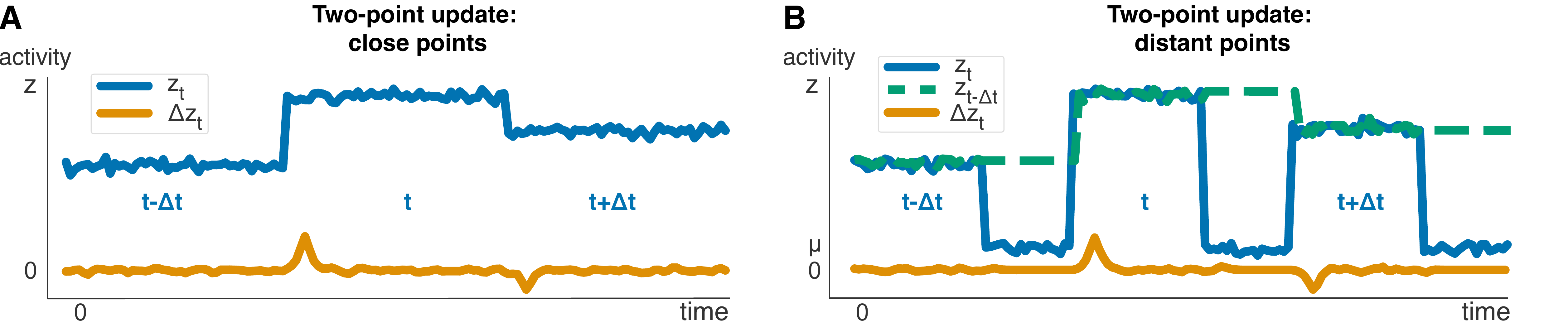}
  \caption{\textbf{A.} First scenario of the Hebbian updates for two points: plasticity (proportional to $\Delta z_t$, orange line) happens when the activity (blue line) switches from one data point to another. \textbf{B.} Second scenario: plasticity happens when the second point comes in some time after the first one, which uses memorized activity from the first point (dashed green line).}
  \label{app:fig:streamed_updates}
\end{figure}
In \cref{sec:online_updates} we proposed online versions for our update rules. For the plain Gaussian kernel (the discussion below also applies to the version with grouping and divisive normalization), the update is
\begin{align}
    \Delta \bb W^k(t)  \propto M^k_{t,\,t-\Delta t} (\bb z^{k}_t - \bb z^{k}_{t-\Delta t}) (\bb z^{k-1}_t - \bb z^{k-1}_{t-\Delta t})\trans\,,
\end{align}
where $t-\Delta t$ represents the data point before the one at time $t$. In the main text we suggested that $z^{k}_{t-\Delta t}$ can be approximated by short-term average of activity in layer $k$.

When $\Delta t$ is small, the change $z^{k}_{n,\,t} - z^{k}_{n,\,t-\Delta t}$ for a neuron $n$ can be computed by a smoothed temporal derivative, implemented by convolution with a kernel $\kappa$ (see \cref{app:fig:streamed_updates}A),
\begin{align}
    z^k_{n,\, t} - z^k_{n,\, t-\Delta t} \approx \Delta z_{n,\, t}^k \equiv (\kappa * z^k_n)(t);\quad \kappa(t)\propto -(t-c_1)\,e^{-c_2|t-c_1|}\,\Theta(t) \, ,
    \label{eq:delta_z_close_points}
\end{align}
with $c_1$ and $c_2$ are positive.

If $\Delta t$ is large and potentially variable, the short-term average won't accurately represent the previous point. However, if between trials $z^k_{n}$ returns to some background activity $\mu^k_n$ (see \cref{app:fig:streamed_updates}B), we can still apply these difference-based updates. In this case the neuron needs to memorize the last significant deviation from the background (i.e., it needs to remember $z^k_{n,\,t-\Delta t} - \mu^k_{n,\,t-\Delta t}$). This can be done with a one-dimensional nonlinear differential equation with ``memory'', such as
\begin{align}
    \dot\omega_{n,\,t} = \brackets{z^k_{n,\,t} - \mu^k_{n,\,t}}^3 - \tanh\brackets{\abs{z^k_{n,\,t} - \mu^k_{n,\,t}}^3}\, \omega_{n,\,t} - c\,\omega_{n,\,t}
    \label{eq:memory_ode}
\end{align}
with $c$ small.
The intuition is as follows: if the neuron is in the background state, the right hand side of \cref{eq:memory_ode} is nearly zero (expect for the leak term $c\,\omega_{n,\,t}$). Otherwise, the deviation from the mean is large and the right hand side approaches $(z^k_{n,\,t} - \mu^k_{n,\,t})^3 - (1+c)\,\omega_{n,\,t}$, as long as $|z^k - \mu^k_{n,\,t}| \gg 1$ (due to tanh saturation). This quickly erases the previous $z^k_{n,\,t-\Delta t}$ and memorizes the new one (see \cref{app:fig:streamed_updates}B with no leak term).
We can therefore compute the difference between the last and the current large deviations by convolving the (real) cube root $(\omega_{n,\,t}^k)^{1/3}$ with the same kernel as above, leading to
\begin{align}
    \bb z^k_{n,\,t} - \bb z^k_{n,\,t-\Delta t} \approx \Delta z^k_{n,\,t}\equiv \big(\kappa * (\omega^k_n)^{1/3}\big)(t)\,.
    \label{eq:delta_z_distant_points}
\end{align}

\subsection{3rd factor for the Gaussian kernel}
The update equation for the Gaussian kernel  (repeating \cref{eq:gauss_update_ij} but in time rather that indices $ij$, and assuming a linear network as it doesn't affect the third factor) is
\begin{align}
    \Delta  W^k_{nm,\, t}  \,&\propto M^k_{t}\, (z^{k}_{n,\,t} - z^{k}_{n,\,t-\Delta t}) (z^{k-1}_{m,\,t} - z^{k-1}_{m,\,t-\Delta t})\,,\label{app:eq:gaussian_update_time}\\
    M^k_{t} \,&= -\frac{1}{\sigma^2}\brackets{ \gamma\,\boldring{k}(\bb y_t, \bb y_{t-\Delta t}) -2\, \boldring{k}(\bb z^{k}_t,\bb z^{k}_{t-\Delta t})}\,k(\bb z^{k}_t,\bb z^{k}_{t-\Delta t})\,.
\end{align}


We'll assume that information about the labels, $k(\bb y_t,\bb y_{t-\Delta t})$, comes from outside the circuit, so to compute the third factor we just need to compute $k(\bb z^{k}_t,\bb z^{k}_{t-\Delta t})$ (and then center everything).
For the Gaussian kernel, we thus need $\norm{\Delta \bb z^k_t}^2$, which is given by
\begin{equation}
    b^k_{1,\,t} = \sum_n (\Delta z^k_{n,\,t})^2\,,
    \label{app:eq:gauss_circuitry_b1}
\end{equation}
so that $k(\bb z^{k}_t,\bb z^{k}_{t-\Delta t}) = \exp(-b_{1,\,t}^k / (2\sigma^2))$.
That gives us the uncentered component of the third factor, denoted $b_{2,\,t}^k\,$,
\begin{align}
\label{app:eq:gauss_circuitry_b2}
    b_{2,\,t}^k \,= \gamma\,{k}(\bb y_t, \bb y_{t-\Delta t}) -2\, {k}(\bb z^{k}_t,\bb z^{k}_{t-\Delta t})
    =\gamma\,{k}(\bb y_t, \bb y_{t-\Delta t}) -2\, \exp\brackets{-\frac{1}{2\sigma^2}b_{1,\,t}^k} \, .
\end{align}
To compute the mean, so that we may center the third factor, we take an exponentially decaying running average,
\begin{align}
    \label{app:eq:gauss_circuitry_b3}   
    b_{3,\,t}^k \,&= \beta\,b_{2,\,t}^k + (1-\beta)\,b_{3,\,t}^k
\end{align}
with $\beta\in(0, 1)$ (so that past points are erased as the weights change).
Thus, the third factor becomes
\begin{align}
    M^k_{t} \,&=-\frac{1}{\sigma^2} \brackets{b^k_{2,\,t} - b^k_{3,\,t}} \exp\brackets{-\frac{1}{2\sigma^2}b_{1,\,t}^k}.
\end{align}

If we think of $b^k_{1,\,t}$, $b^k_{2,\,t}$ and $b^k_{3,\,t}$ as neurons, the first two need nonlinear dendrites to compute this signal. In addition, $b^k_{1,\,t}$ should compute $\Delta z^k_{n,\,t}$ from $z^k_{n,\,t}$ at the dendritic level.

\subsection{3rd factor for the  Gaussian kernel with grouping and divisive normalization}
The update for the Gaussian kernel with grouping (repeating \cref{eq:divnorm_update} but in time, and assuming a linear network as it doesn't affect the third factor) is
\begin{subequations}
\begin{align}
\label{hebbian_grouping}
    \Delta W^k_{\alpha n m,\, t}  \,&\propto M_{\alpha,\,t}^k \brackets{r^{k}_{\alpha n,\,t} r^{k-1}_{m,\,t} - r^{k}_{\alpha n,\,t-\Delta t} r^{k-1}_{m,\,t-\Delta t}}\,,\\
    M_{\alpha,\,t}^k\,&=M_{t}^k\,({v}_{\alpha,\,t}^k - {v}_{\alpha,\,t-\Delta t}^k)\,,\\
    M^k_{t} \,&= -\frac{1}{\sigma^2}\brackets{ \gamma\,\boldring{k}(\bb y_t, \bb y_{t-\Delta t}) -2\, \boldring{k}(\bb z^{k}_t,\bb z^{k}_{t-\Delta t})}\,k(\bb z^{k}_t,\bb z^{k}_{t-\Delta t})\,.
\end{align}
\end{subequations}
The Hebbian term -- the term in parentheses in \cref{hebbian_grouping} -- corresponds to the difference of pre- and post-synaptic products, rather than a product of differences:  $r^{k}_{\alpha n,\,t} r^{k-1}_{m,\,t} - r^{k}_{\alpha n,\,t-\Delta t} r^{k-1}_{m,\,t-\Delta t}\approx \Delta(r^{k}_{\alpha n,\,t} r^{k-1}_{m,\,t})$. We can compute this difference as before (\cref{eq:delta_z_close_points} or \cref{eq:delta_z_distant_points}), although this update rule requires a different interaction of the pre- and post-synaptic activity comparing to the plain Gaussian kernel in \cref{app:eq:gaussian_update_time}.

The third factor is almost the same as for the plain Gaussian kernel, but there are two differences.
First, we need to compute the centered normalization $v_{\alpha,\,t}^k$ (as in \cref{eq:grouping_v}) and its change over time for each group $\alpha$,
\begin{align}
    \tilde b_{\alpha,\,t}^k = \Delta v_{\alpha,\,t}^k\,.
\end{align}

Second, $M_t^k$ is computed for $\Delta v^k_{\alpha,\, t}$ rather than $\Delta z^k_{n,\, t}$, such that (cf. \cref{app:eq:gauss_circuitry_b1})
\begin{align}
    \tilde b^k_{1,\,t} \,&= \sum_\alpha (\tilde b_{\alpha,\,t}^k)^2\,,
\end{align}
and $\tilde b^k_{2,\, t}$ (\cref{app:eq:gauss_circuitry_b2} but with $\tilde b_{1,\,t}^k$) and $\tilde b^k_{3,\, t}$ (\cref{app:eq:gauss_circuitry_b3} but with $\tilde b_{1,\,t}^k$ and $\tilde b_{2,\,t}^k$) stay the same.

The third factor becomes
\begin{align}
    M^k_{\alpha,\,t} \,&=-\frac{1}{\sigma^2} \brackets{\tilde b^k_{2,\,t} - \tilde b^k_{3,\,t}} \exp\brackets{-\frac{1}{2\sigma^2}\tilde b_{1,\,t}^k}\tilde b_{\alpha,\,t}^k\,.
\end{align}
Essentially, it is the same third factor computation as for the Gaussian kernel, but with an additional group-specific signal $b_{\alpha,\,t}^k$.

\section{Experimental details}
\label{app:sec:experiments}

\subsection{Network architecture}
Each hidden layer of the network had its own optimizer, such that weight updates happen during the forward pass. 

Each layer had the following structure: linear/convolutional operation $\rightarrow$ batchnorm (if any) $\rightarrow$ nonlinearity $\rightarrow$ pooling (if any) $\rightarrow$ local loss computation (doesn't modify activity) $\rightarrow$ divisive normalization (if any) $\rightarrow$ dropout.

None of the hidden layers had the bias term, but the output layer did. The last layer (or the whole network for backprop) was trained with the cross-entropy loss. Non-pHSIC methods with divisive normalization used the same group arrangement, but did not have grouping in the objectives.

\subsection{Choice of kernels for pHSIC}
The gradient over each batch was computed as in \cref{app:eq:phsic_obj_deriv}. 

We used cosine similarity with centered labels (\cref{eq:y_kernel_balanced}; as all datasets are balanced, we don't need to know the probability of a class to center). The kernels for $\bb z$ were plain Gaussian (\cref{app:eq:gaussian_kernel}), Gaussian with grouping and divisive normalization (``grp+div''; \cref{grouped_kernel} such that the next layer sees $r_{\alpha n}^k\equiv \boldring z_{\alpha n}^k / (u_{\alpha}^k)^{p}$) and grouping without divisive normalization (``grp''; also \cref{grouped_kernel} but the next layer sees $z_{\alpha n}^k$), plain cosine similarity (\cref{app:eq:cossim_kernel}), and cosine similarity with grouping and with or without divisive normalization ($k(\bb z^k_i,\bb z^k_j) = (\bb v_i^k)\trans\bb v_j^k / (\norm{\bb v_i^k}\norm{\bb v_j^k})$ with $\bb v$ as in \cref{eq:grouping_v}).

\subsection{Objective choice for layer-wise classification}
As proposed in \cite{nokland2019training}, each convolutional layer is first passed through an average pooling layer such that the final number of outputs is equal to 2048 (e.g. a layer with 128 channels and 32 by 32 images is pooled with an 8 by 8 kernel with stride$\,=8$); the resulting 2048-dimensional vector is transformed into a 10-dimensional vector (for class prediction) by a linear readout layer. Fully connected layers are transformed directly with the corresponding linear readout layer. In the layer-wise classification with feedback alignment, feedback alignment is applied to the readout layer.

\subsection{Pre-processing of datasets} 
MNIST, fashion-MNIST and Kuzushiji-MNIST images were centered by $0.5$ and normalized by $0.5$.

For CIFAR10, each training image was padded by zeroes from all sides with width 4 (resulting in a 40 by 40 image for each channel) and randomly cropped to the standard size (32 by 32), then flipped horizontally with probability $0.5$, and then centered by $(0.4914, 0.4822, 0.4465)$ (each number corresponds to a channel) and normalized by $(0.247, 0.243, 0.261)$. For validation and test, the images were only centered and normalized.

\subsection{Shared hyperparameters for all experiments}
We used the default parameters for AdamW, batchnorm, LReLU and SELU; for grouping without divisive normalization we used $p=0.5$ to be comparable with the objective in \cite{nokland2019training}. The rest of the parameters (including the ones below) were tuned on a validation set ($10\%$ of the training set for all datasets).

Weight decay for the local losses was 1e-7, and for the final/backprop it was 1e-6; the learning rates were multiplied by $0.25$, with individual schedules described below. For SGD, the momentum was 0.95; for AdamW, $\beta_1=0.9$, $\beta_2=0.999$, $\eps=$1e-8. Batchnorm had momentum of 0.1 and $\eps=1e-5$, with initial scale $\gamma=1$ and shift $\beta=0$. Leaky ReLU had the slope 0.01; $\mathrm{SELU}(x)=\textrm{scale}(\max(0,x)+\min(0,\alpha (\exp(x) - 1)))$ had
$\alpha \approx 1.6733$ and $\textrm{scale} \approx 1.0507$ (precise values were found numerically in \cite{klambauer2017self}; note that dropout for SELU was changed to alpha dropout, as proposed in \cite{klambauer2017self}). All convolutions used 3 by 3 kernel with padding$\,=1$ (on each side), stride$\,=1$ and dilation$\,=1$ and no groups; max pooling layers used 2 by 2 kernels with stride$\,=2$, dilation$\,=1$ and no padding. Grouping with divisive normalization used $\delta=1$ and $p=0.2$ (backprop, pHSIC) or $p=0.5$ (FA, sign symmetry, layer-wise classification), and $p=0.5$ without divisive normalization. Gaussian kernels used $\sigma=5$. The balance parameter was set to $\gamma=2$.

\subsection{Small network}
The dropout for all experiments was 0.01, with LReLU for nonlinearity. The networks were trained for 100 epochs, and the learning rates were multiplied by 0.25 at epochs 50, 75 and 90. The individual parameters, $\eta_f$ for final/backprop initial learning rate, $\eta_l$ for the local initial learning rate, $c^k$ for the number of groups in the objective, are given in \cref{app:tab:results:mlp_params}; the final results are given in \cref{app:tab:results:mlp_full} (same as \cref{tab:results:mlp} but with ``grp'') and \cref{app:tab:results:mlp_minmax} (max - min accuracy).

\subsection{Large network}
The dropout for all experiments was 0.05, with LReLU for AdamW+batchnorm and SELU for SGD. The networks were trained for 500 epochs, and the learning rates were multiplied by 0.25 at epochs 300, 350, 450 and 475 (and at 100, 200, 250, 275 for backprop with SGD). The individual parameters, $\eta_f$ for final/backprop initial learning rate, $\eta_l$ for the local initial learning rate, $c^k$ for the number of groups in the objective, are given in \cref{app:tab:results:vgg_params} and \cref{app:tab:results:vgg_params_new}. The results are given in \cref{app:tab:results:vgg_full} and \cref{app:tab:results:vgg_full_new} (mean test accuracy) and \cref{app:tab:results:vgg_full_minmix} and \cref{app:tab:results:vgg_full_minmax_new} (max - min accuracy). The batch manhattan method mentioned in \cref{app:tab:results:vgg_params_new} was proposed in \cite{liao2015important}; it is used to stabilize feedback alignment and sign symmetry algorithms by substituting the gradient w.r.t. the loss (i.e. before adding momentum and weight decay) with its sign for each weight update. However, in our experiments it didn't improve performance in most of the cases.
Without any normalization, we didn't find a successful set of parameters for the Gaussian kernel with grouping and for the methods with feedback alignment and sign symmetry. In those cases, the training either diverged completely or was stuck at low training and even lower test errors (e.g. around 40\% training error for the Gaussian kernel with grouping, and around 80\% training error for layer-wise classification with feedback alignment).

\subsection{Difference between pHSIC and HSIC in the large network}
While we reported all result for pHSIC, training with HSIC instead did not lead to a significant change in the results (not shown). Moreover, the difference between the two objectives stays small during training, as we illustrate below.

As explained in \cref{app:subsec:phsic}, our objectives differs from HSIC only in the first term, $\mathrm{p}\hsic{Z^k}{Z^k}$, due to centering of labels. We trained the 1x wide networks from the previous section (grouping + divisive normalization with SGD, grouping + batchnorm with AdamW) and plotted
\begin{equation}
    \frac{\mathrm{p}\hsic{Z^k}{Z^k} - \hsic{Z^k}{Z^k}}{\mathrm{p}\hsic{Z^k}{Z^k}}\,
    \label{app:eq:hsic_rel_distance}
\end{equation}
as a function of training epoch. We compute this quantity on the training data, but the test data gives the same behavior (not shown).

\begin{figure}[h]
  \centering
  \includegraphics[width=\textwidth]{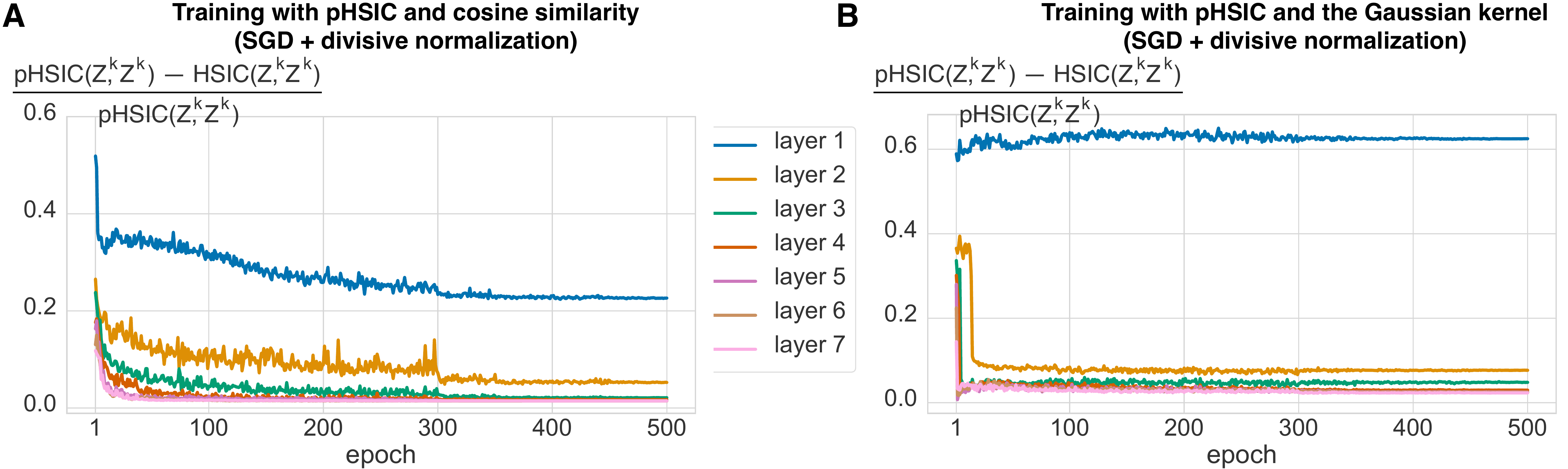}
  \caption{Training of 1x networks with pHSIC, SGD and divisive normalization. Y-axis represents $\brackets{\mathrm{pHSIC}(Z^k, Z^k) - \mathrm{HSIC}(Z^k, Z^k)} / \mathrm{pHSIC}(Z^k, Z^k)$. \textbf{A.} Cosine similarity kernel \textbf{B.} Gaussian kernel.}
  \label{app:fig:phsic_sgd}
\end{figure}

\begin{figure}[h]
  \centering
  \includegraphics[width=\textwidth]{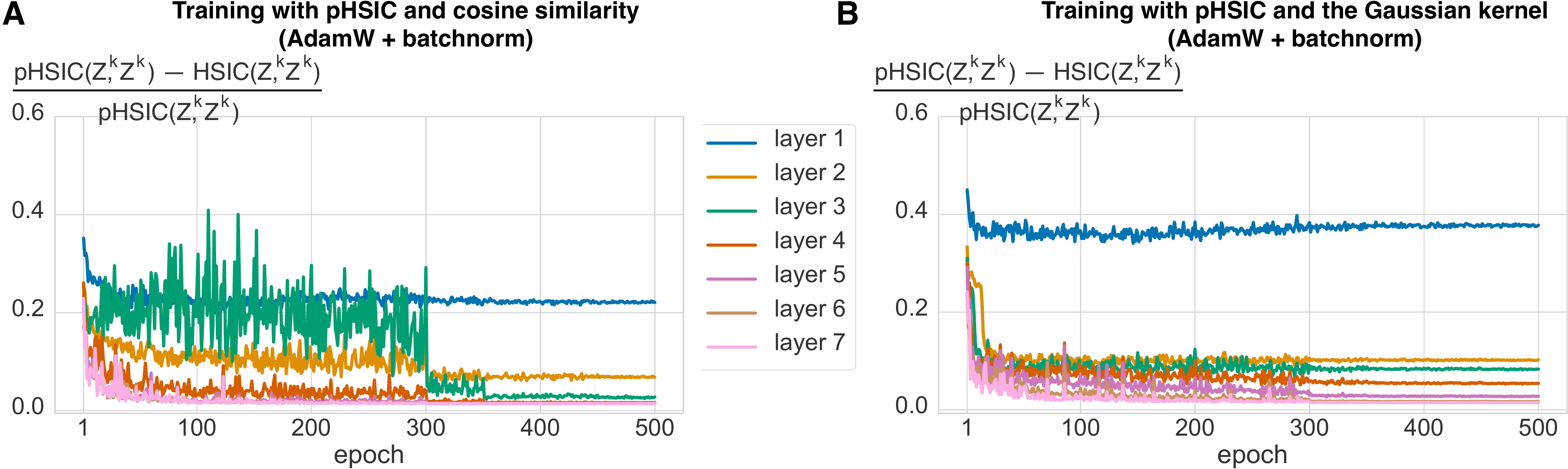}
  \caption{Training of 1x networks with pHSIC, AdamW and batchnorm. Y-axis represents $\brackets{\mathrm{pHSIC}(Z^k, Z^k) - \mathrm{HSIC}(Z^k, Z^k)} / \mathrm{pHSIC}(Z^k, Z^k)$. \textbf{A.} Cosine similarity kernel \textbf{B.} Gaussian kernel.}
  \label{app:fig:phsic_adam}
\end{figure}

\begin{figure}[h]
  \centering
  \includegraphics[width=\textwidth]{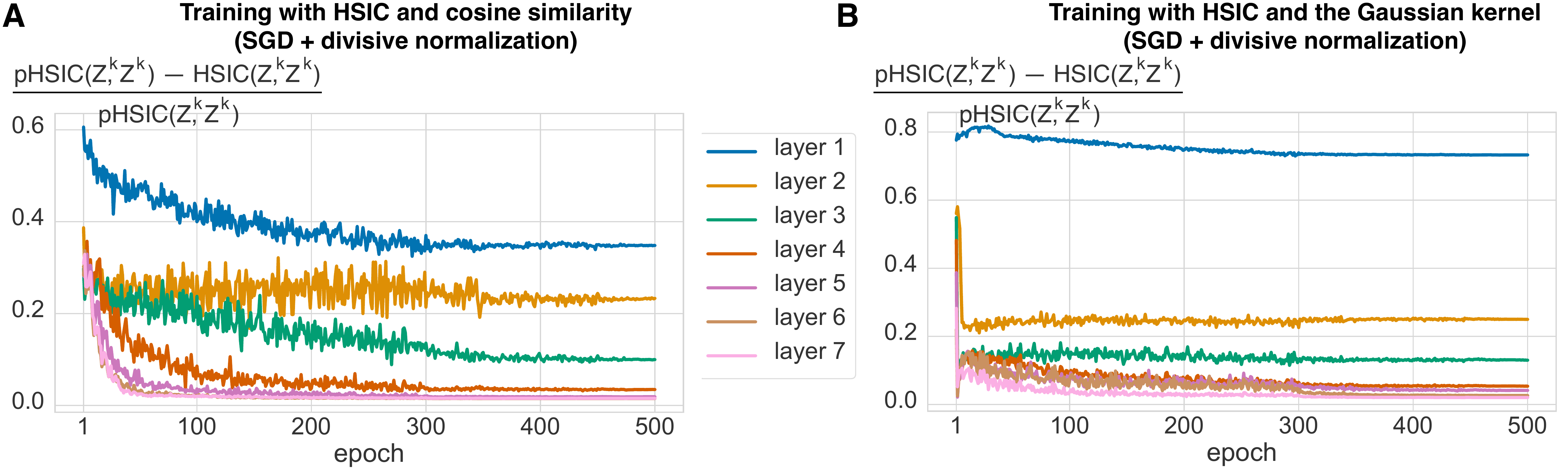}
  \caption{Training of 1x networks with HSIC, SGD and divisive normalization. Y-axis represents $\brackets{\mathrm{pHSIC}(Z^k, Z^k) - \mathrm{HSIC}(Z^k, Z^k)} / \mathrm{pHSIC}(Z^k, Z^k)$. \textbf{A.} Cosine similarity kernel \textbf{B.} Gaussian kernel.}
  \label{app:fig:hsic_sgd}
\end{figure}

\begin{figure}[h]
  \centering
  \includegraphics[width=\textwidth]{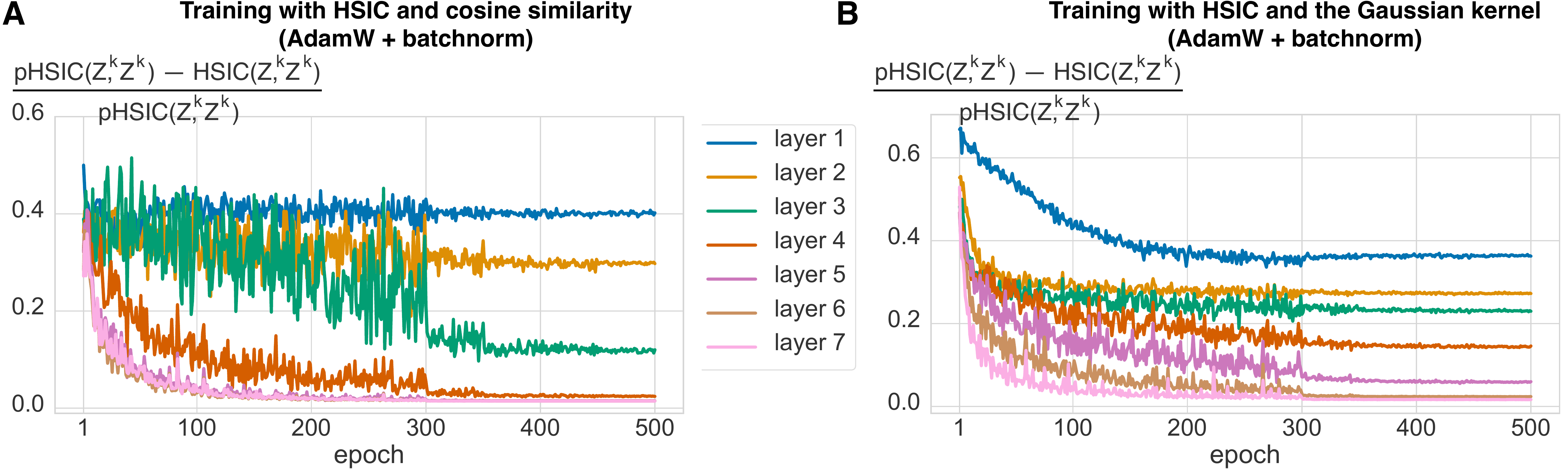}
  \caption{Training of 1x networks with HSIC, AdamW and batchnorm. Y-axis represents $\brackets{\mathrm{pHSIC}(Z^k, Z^k) - \mathrm{HSIC}(Z^k, Z^k)} / \mathrm{pHSIC}(Z^k, Z^k)$. \textbf{A.} Cosine similarity kernel \textbf{B.} Gaussian kernel.}
  \label{app:fig:hsic_adam}
\end{figure}

The results show that for both the cosine similarity and the Gaussian kernel, the relative distance between pHSIC and HSIC (\cref{app:eq:hsic_rel_distance}) stays small in all layers expect the first one, but even there it remains relatively constant when trained on the pHSIC objective with SGD + divisive normalization (\cref{app:fig:phsic_sgd}) or AdamW + batchnorm (\cref{app:fig:phsic_adam}); the same holds when the objective is HSIC (\cref{app:fig:hsic_sgd} for SGD + divisive normalization and \cref{app:fig:hsic_adam} for AdamW + batchnorm), although earlier layers have larger values when compared to pHSIC training.

\begin{table}[h]
  \caption{Parameters for the 3-layer fully connected net (1024 neurons per layer). Last layer: training of the last layer; cossim: cosine similarity; grp: grouping; div: divisive normalization. 
  }
  \label{app:tab:results:mlp_params}
  \centering
  \begin{tabular}{rcccccccccc}
    \toprule
    & \multicolumn{2}{c}{backprop} & \multicolumn{2}{c}{last layer} & \multicolumn{3}{c}{pHSIC: cossim} & \multicolumn{3}{c}{pHSIC: Gaussian}\\ \cmidrule(r){2-3}\cmidrule(r){4-5}\cmidrule(r){6-8}\cmidrule(r){9-11}& & div & & div & & grp & grp+div  & & grp & grp+div \\ \midrule MNIST\\
    $\eta_f$ & 5e-2 & 5e-3 & 5e-2 & 5e-2 & 5e-3 & 5e-3 & 5e-3 & 5e-4 & 5e-4 & 1e-3 \\
    $\eta_l$ & & & & & 0.5 & 0.6 & 0.4 & 0.6 & 1.0& 1.0\\
    $c^k$ & & 16 & & 16 & & 16 & 16 &  & 32 & 32 \\
    f-MNIST\\
    $\eta_f$ & 5e-3 & 5e-3  & 5e-2 & 5e-2  & 5e-3 & 1e-3 & 5e-4    & 5e-4 & 5e-4 & 5e-4 \\
    $\eta_l$ & &   & &   & 1.0 & 0.6 & 1.0  & 0.5 & 1.0 & 1.0 \\
    $c^k$    & & 32  & & 32   & & 32 & 32    & & 32 & 32 \\
    K-MNIST\\
    $\eta_f$ & 5e-2 & 5e-2  & 5e-2 & 5e-2  & 5e-3 & 5e-3 & 5e-4    & 1e-3 & 1e-3 & 1e-3\\
    $\eta_l$ & &   & &   & 0.6 & 0.4 & 0.4   & 0.6 & 1.0 & 1.0 \\
    $c^k$    & & 32   & & 16  & & 16 & 16    & & 32 & 32\\
    CIFAR10\\
    $\eta_f$ & 5e-3 & 5e-3   & 5e-2 & 1e-2   & 1e-3 & 5e-3 & 5e-3   & 5e-3 & 5e-4 & 1e-3 \\
    $\eta_l$ & & 32   & & 32   & 1.0 & 0.4 & 0.1    & 0.1 & 0.6 & 1.0 \\
    $c^k$    & &   & &   & & 32 & 32   & & 32 & 32\\
    \bottomrule
  \end{tabular}
\end{table}

\begin{table}[h]
  \caption{Mean test accuracy over 5 random seeds for a 3-layer fully connected net (1024 neurons per layer). Last layer: training of the last layer; cossim: cosine similarity; grp: grouping; div: divisive normalization.}
  \label{app:tab:results:mlp_full}
  \centering
  \begin{tabular}{lcccccccccc}
    \toprule
    & \multicolumn{2}{c}{backprop} & \multicolumn{2}{c}{last layer} & \multicolumn{3}{c}{pHSIC: cossim} & \multicolumn{3}{c}{pHSIC: Gaussian}\\ \cmidrule(r){2-3}\cmidrule(r){4-5}\cmidrule(r){6-8}\cmidrule(r){9-11}& & grp+div & & grp+div & & grp & grp+div  & & grp & grp+div \\ \midrule MNIST & 98.6& 98.4& 92.0& 95.4& 94.9& 95.8& 96.3& 94.6& 98.4& 98.1\\f-MNIST & 90.2& 90.8& 83.3& 85.7& 86.3& 88.7& 88.1& 86.5& 88.6& 88.8\\K-MNIST & 93.4& 93.5& 71.2& 78.2& 80.4& 86.2& 87.2& 80.2& 92.7& 91.1\\CIFAR10 & 60.0& 60.3& 39.2& 38.0& 51.1& 52.5& 47.6& 41.4& 48.4& 46.4\\
    \bottomrule
  \end{tabular}
\end{table}

\begin{table}[h]
  \caption{Same as \cref{app:tab:results:mlp_full}, but max minus min test accuracy over 5 random seeds.}
  \label{app:tab:results:mlp_minmax}
  \centering
  \begin{tabular}{lcccccccccc}
    \toprule
    & \multicolumn{2}{c}{backprop} & \multicolumn{2}{c}{last layer} & \multicolumn{3}{c}{pHSIC: cossim} & \multicolumn{3}{c}{pHSIC: Gaussian}\\ \cmidrule(r){2-3}\cmidrule(r){4-5}\cmidrule(r){6-8}\cmidrule(r){9-11}& & grp+div & & grp+div & & grp & grp+div  & & grp & grp+div \\ \midrule MNIST & 0.2& 0.1& 0.3& 0.3& 1.4& 0.5& 0.6& 0.2& 0.3& 0.2\\f-MNIST & 0.2& 0.4& 0.3& 0.2& 0.6& 1.1& 0.3& 0.2& 0.6& 0.2\\K-MNIST & 0.3& 0.3& 1.1& 0.8& 1.0& 1.0& 0.9& 1.0& 0.4& 1.2\\CIFAR10 & 0.6& 0.9& 1.2& 1.4& 1.4& 2.0& 1.4& 0.5& 1.0& 0.6\\
    \bottomrule
  \end{tabular}
\end{table}

\begin{table}[h]
  \caption{Parameters for the 7-layer conv nets (CIFAR10; 1x and 2x wide). Cossim: cosine similarity; divnorm: divisive normalization; bn: batchnorm. Empty entries: experiments for which we didn't find a satisfying set of parameters due to instabilities in the methods.}
  \label{app:tab:results:vgg_params}
  \centering
  \begin{tabular}{rcccccccccc}
    \toprule
    & \multicolumn{2}{c}{backprop} & \multicolumn{2}{c}{pHSIC: cossim} & \multicolumn{2}{c}{pHSIC: Gaussian}\\ \cmidrule(r){2-3}\cmidrule(r){4-5}\cmidrule(r){6-7} & & div & grp & grp+div & grp & grp+div \\ \midrule
    1x wide net + SGD\\
    $\eta_f$ & 5e-3 & 6e-3    & 5e-5 & 5e-4   & & 1e-4 \\
    $\eta_l$ & &    & 3e-2 & 0.5  & & 0.4\\
    $c^k$    & & 64    & 32 & 64   & & 64 \\
    2x wide net + SGD\\
    $\eta_f$ & 6e-3 & 6e-3    & 5e-5 & 5e-4  & & 1e-4 \\
    $\eta_l$ & &    & 3e-2 &  0.5  & & 0.4 \\
    $c^k$    & & 64   & 32 & 64  & & 64 \\
    1x wide net + AdamW + batchnorm\\
    $\eta_f$ & 5e-3 & 5e-3   & 5e-4 & 5e-4   & 5e-4 & 5e-4\\
    $\eta_l$ & &    & 5e-4 & 5e-4   & 5e-3 & 1e-2 \\
    $c^k$    &  & 64    & 32 & 64  & 64 & 64 \\
    2x wide net + AdamW + batchnorm\\
    $\eta_f$ & 5e-3 & 5e-3    & 5e-4 & 5e-4   & 5e-4 & 5e-4 \\
    $\eta_l$ & &    & 5e-4 & 5e-4   & 5e-3 & 5e-3\\
    $c^k$    & & 64   & 128 & 64  & 128 & 128 \\
    \bottomrule
  \end{tabular}
\end{table}

\begin{table}[h]
  \caption{Parameters for the 7-layer conv nets (CIFAR10; 1x and 2x wide). FA: feedback alignment; sign sym.: sign symmetry; layer class.: layer-wise classification; divnorm: divisive normalization; bn: batchnorm. Empty entries: experiments for which we didn't find a satisfying set of parameters due to instabilities in the methods.}
  \label{app:tab:results:vgg_params_new}
  \centering
  \begin{tabular}{rcccc}
    \toprule
    & FA & sign sym. & \multicolumn{2}{c}{layer class.}\\\cmidrule(r){4-5}
    &    &           & & +FA \\ \midrule
    1x wide net + SGD\\
                $\eta_f$        & & & 5e-3 & \\
                $\eta_l$        & & & 1e-3 & \\
    2x wide net + SGD\\
                    $\eta_f$    & & & 5e-3 & \\
                $\eta_l$        & & & 1e-3 & \\
    1x wide net + SGD + divnorm\\
                    $\eta_f$    & 1e-3 & 5e-4 & 5e-3 & 5e-3 \\
                $\eta_l$        &      & & 5e-3 & 5e-3 \\
                $c^k$        & 64 & 64 & 64 & 64 \\
                batch manhattan        & & + & & \\
    2x wide net + SGD + divnorm \\
                    $\eta_f$ & 5e-4 & 5e-4 & 5e-3 & 5e-3 \\
                $\eta_l$     & & & 5e-3 & 5e-3 \\
                $c^k$        & 128 & 128 & 64 & 128\\
                batch manhattan        & & + & & \\
    1x wide net + AdamW + bn \\
                    $\eta_f$    & 5e-4 & 5e-4 & 5e-3 & 5e-3 \\
                $\eta_l$        &   & & 5e-4 & 1e-3 \\
    2x wide net + AdamW + bn\\
                    $\eta_f$    & 5e-4 & 5e-4 & 5e-3 & 5e-3\\
                $\eta_l$        & & & 5e-4 & 5e-4\\
    \bottomrule
  \end{tabular}
\end{table}

\begin{table}[h]
  \caption{Mean test accuracy on CIFAR10 over 5 runs for a 7-layer conv nets (1x and 2x wide). Cossim: cosine similarity; divnorm: divisive normalization; bn: batchnorm. Empty entries: experiments for which we didn't find a satisfying set of parameters due to instabilities in the methods.}
  \label{app:tab:results:vgg_full}
  \centering
  \begin{tabular}{lcccccccccc}
    \toprule
    & \multicolumn{2}{c}{backprop} & \multicolumn{2}{c}{pHSIC: cossim} & \multicolumn{2}{c}{pHSIC: Gaussian}\\ \cmidrule(r){2-3}\cmidrule(r){4-5}\cmidrule(r){6-7} & & div & grp & grp+div & grp & grp+div \\ \midrule
    1x wide net + SGD & 91.0 & 91.0 & 88.8 & 89.8  &  & 86.2\\
    2x wide net + SGD & 91.9 & 90.9 & 89.4 & 91.3  &  &  90.4 \\
    1x wide net + AdamW + batchnorm & 94.1 & 94.3 & 91.3 & 90.1  & 89.9 & 89.4 \\
    2x wide net + AdamW + batchnorm & 94.3 & 94.5 & 91.9 & 91.0  & 91.0 &  91.2 \\
    \bottomrule
  \end{tabular}
\end{table}

\begin{table}[h]
  \caption{Mean test accuracy on CIFAR10 over 5 runs for a 7-layer conv nets (1x and 2x wide). FA: feedback alignment; sign sym.: sign symmetry; layer class.: layer-wise classification; divnorm: divisive normalization; bn: batchnorm. Empty entries: experiments for which we didn't find a satisfying set of parameters due to instabilities in the methods.}
  \label{app:tab:results:vgg_full_new}
  \centering
  \begin{tabular}{lcccc}
    \toprule
    & FA & sign sym. & \multicolumn{2}{c}{layer class.}\\\cmidrule(r){4-5}
    &    &           & & +FA \\ \midrule
    1x wide net + SGD           & & & 90.0 & \\
    2x wide net + SGD           & & & 90.3 &\\
    1x wide net + SGD + divnorm & 80.4 & 89.5 & 90.5 & 81.0\\
    2x wide net + SGD + divnorm & 80.6 & 91.3 & 91.3 & 81.2\\
    1x wide net + AdamW + bn    & 82.4 & 93.6 & 92.1 & 90.3 \\
    2x wide net + AdamW + bn    & 81.6 & 93.9 & 92.1 & 91.1  \\
    \bottomrule
  \end{tabular}
\end{table}

\begin{table}[h]
  \caption{Same as \cref{app:tab:results:vgg_full}, but max minus min test accuracy over 5 random seeds.}
  \label{app:tab:results:vgg_full_minmix}
  \centering
  \begin{tabular}{lcccccccccc}
    \toprule
    & \multicolumn{2}{c}{backprop} & \multicolumn{2}{c}{pHSIC: cossim} & \multicolumn{2}{c}{pHSIC: Gaussian}\\ \cmidrule(r){2-3}\cmidrule(r){4-5}\cmidrule(r){6-7} & & div & grp & grp+div & grp & grp+div \\ \midrule
    1x wide net + SGD& 0.4 & 0.4 & 0.7 & 0.7  &  & 0.9\\
    2x wide net + SGD& 0.3 & 0.3 & 2.4 & 0.2  &  &  0.5 \\
    1x wide net + AdamW + batchnorm & 0.3 & 0.4 & 0.2 & 0.5  & 0.3 & 0.5 \\
    2x wide net + AdamW + batchnorm & 0.5 & 0.3 & 0.3 & 0.3  & 0.4 &  0.5 \\
    \bottomrule
  \end{tabular}
\end{table}

\begin{table}[h]
  \caption{Same as \cref{app:tab:results:vgg_full_new}, but max minus min test accuracy over 5 random seeds. *The large deviation is due to one experiment with about 85\% accuracy.}
  \label{app:tab:results:vgg_full_minmax_new}
  \centering
  \begin{tabular}{lcccc}
    \toprule
    & FA & sign sym. & \multicolumn{2}{c}{layer class.}\\\cmidrule(r){4-5}
    &    &           & & +FA \\ \midrule
    1x wide net + SGD           & & & 0.4 &\\
    2x wide net + SGD           & & & 0.1 &\\
    1x wide net + SGD + divnorm & 1.3 & 5.5* & 0.4 & 1.1 \\
    2x wide net + SGD + divnorm & 0.9 & 0.4 & 0.1 & 0.9\\
    1x wide net + AdamW + bn    & 0.4 & 0.3 & 0.6 & 0.5\\
    2x wide net + AdamW + bn    & 0.9 & 0.4 & 0.1 &0.4 \\
    \bottomrule
  \end{tabular}
\end{table}



\end{document}